\documentclass[11pt]{article}
\usepackage[
lmargin=3cm,rmargin=3cm,
textheight=8.9in,
top=1in,
]{geometry}
\usepackage[utf8]{inputenc}
\usepackage[T1]{fontenc}

\usepackage{amsmath,amsthm,amsfonts}
\usepackage{enumitem}
\usepackage[makeroom]{cancel}
\usepackage{graphicx}
\usepackage{tabularx}
\usepackage{booktabs}
\usepackage{caption}
\usepackage{subcaption}

\usepackage{natbib}
\setcitestyle{authoryear}
\usepackage{authblk}

\usepackage[noend,linesnumbered,ruled]{algorithm2e}
\usepackage{xcolor}
\usepackage{hyperref}
\definecolor{mydarkblue}{rgb}{0,0.08,0.45}
\hypersetup{ %
pdftitle={},
pdfauthor={},
pdfkeywords={},
colorlinks=true,
linkcolor=mydarkblue,
citecolor=mydarkblue,
filecolor=mydarkblue,
urlcolor=mydarkblue}

\newcommand\Acal{\mathcal A}
\newcommand\Ncal{\mathcal N}
\newcommand\Ccal{\mathcal{C}}
\newcommand\Dcal{\mathcal{D}}
\newcommand\Jcal{\mathcal{J}}
\newcommand\Gcal{\mathcal{G}}
\newcommand\Lcal{\mathcal{L}}
\newcommand\Ocal{\mathcal{O}}
\newcommand\Scal{\mathcal{S}}

\newcommand{\Rbb}{\mathbb{R}}

\newcommand{\proj}{\Pi}

\newcommand{\xpub}{x^{\text{pub}}}
\newcommand{\xpriv}{x^{\text{priv}}}
\newcommand{\Xpub}{X^{\text{pub}}}

\newcommand\Oj{\Omega_j}
\newcommand\Ok{\Omega^k}
\newcommand\Exp{\mathbb E}
\newcommand\datasize{D}
\newcommand\clip{\text{Clip}}
\DeclareMathOperator{\argmin}{argmin}

\newcommand\userUpdate{\texttt{UserUpdate}}
\newcommand\itemUpdate{\texttt{ItemUpdate}}
\newcommand\blue[1]{{\color{blue} #1}}

\newtheorem{assump}[]{Assumption}
\newtheorem{lemma}[]{Lemma}
\newtheorem{proposition}[]{Proposition}
\newtheorem{remark}[]{Remark}
\newtheorem{definition}[]{Definition}
\newtheorem{theorem}[]{Theorem}

\title{Private Learning with Public Features}

\date{}
\author{Walid Krichene}
\author{Nicolas Mayoraz}
\author{Steffen Rendle}
\author{\\Shuang Song}
\author{Abhradeep Thakurta}
\author{Li Zhang\footnote{Work done while at Google.}}

\affil{Google}

\begin{document}

\maketitle
\begin{abstract}
We study a class of private learning problems in which the data is a join of \emph{private and public features}. This is often the case in private personalization tasks such as recommendation or ad prediction, in which features related to individuals are sensitive, while features related to items (the movies or songs to be recommended, or the ads to be shown to users) are publicly available and do not require protection. A natural question is whether private algorithms can achieve higher utility in the presence of public features.
We give a positive answer for multi-encoder models where one of the encoders operates on public features. We develop new algorithms that take advantage of this separation by only protecting certain sufficient statistics (instead of adding noise to the gradient). This method has a guaranteed utility improvement for linear regression, and importantly, achieves the state of the art on two standard private recommendation benchmarks, demonstrating the importance of methods that adapt to the private-public feature separation.
\end{abstract}

\section{Introduction}

Models trained on private user data risk leaking sensitive information~\citep{korolova2010privacy,calandrino2011you,zhu2019deep}, and Differential Privacy (DP)~\citep{dwork2006calibrating} offers ways to quantify and limit this risk.
One of the main challenges of DP training is that privacy guarantees come at the expense of losses in utility -- the state of the art private models suffer significant quality losses compared to their non-private counterpart in many benchmarks. To mitigate these losses, a large body of work~(e.g., \cite{bassily2020learning,yu2021differentially,zhou2021bypassing,golatkar2022mixed,li2022private,amid2022public,bassily2023principled,ganesh2023why} among others) has explored methods to improve utility by leveraging public data sources. For example, it was shown that in image classification and certain language tasks, pre-training models on large public data can significantly improve utility of private models~\citep{li2022large,de2022unlocking,mehta2023towards,ganesh2023why}.

The vast majority of these approaches assume access to \emph{public training examples}. A different practical setting that remains under-explored--and which we propose to study--is having access to \emph{public features}. This is often the case in personalization problems such as recommendation or ad prediction, where the same training example contains sensitive features about an individual, as well as features about an item that the individual interacts with (e.g., the recommended movie or song, or the advertised product). These item features are public and don't need privacy protection; for example, the director of a movie, the genre of a song, or the brand of a product are all public information. Formally, we assume access to a feature matrix $\Xpub$ that is public, and each training example contains \emph{some row} of this matrix, together with private features and labels. Our goal is to design private algorithms adapted to this setting, and to study whether access to $\Xpub$ can improve privacy/utility trade-offs.

\subsection{Contributions}
\begin{enumerate}[leftmargin=13pt,topsep=0pt,itemsep=0ex,partopsep=0ex,parsep=1pt]
\item We design new first-order algorithms for private learning with public features. The main algorithmic novelty is that instead of protecting gradients (as one would do in noisy gradient descent~\citep{bassily2014erm} and its variants), our algorithms work by computing and protecting certain sufficient statistics, that are then used to compute the final gradient. This has several practical advantages that we will discuss in detail; a notable one is that this allows us to preserve gradient sparsity, addressing a major practical issue of noisy gradient descent.
\item We give a utility analysis for linear regression (Theorem~\ref{thm:utility}), where we compare the utility when some features are public, to the utility when all features are private. We show that the excess empirical risk improves by a factor of $\sqrt{m/p}$ ($m$ is the number of items, and $p$ is the number of public features). In other words, the typical $\sqrt{p}$ scaling is replaced with a $\sqrt{m}$ scaling. To the best of our knowledge, this is the first analysis that shows a utility improvement under public features.
\item Empirically, we evaluate our methods on two standard private recommendation benchmarks (one classification task and one regression task from the MovieLens data sets~\citep{harper16movielens}). Our method achieves the DP state of the art on both benchmarks, demonstrating the importance of algorithms that adapt to the private-public feature structure. Besides improving utility, our method is also more computationally efficient, achieving better quality than DP-SGD at a fraction of the computational cost. We conclude with a discussion of limitations and open questions.
\end{enumerate}

%offer practical advantages such as efficient retrieval using maximum inner-product search~\citep{shrivastava2014mips,guo2020scann}.
\subsection{Related Work}

\paragraph{Private learning with public data} Several approaches have been proposed to augment private training with public data. Existing methods broadly fall into two categories. The first is to use the public data to modify the training procedure: a common theme is to project gradients on a low-dimensional manifold estimated using public data~\citep{kairouz2020fast,yu2021large,zhou2021bypassing}, or to augment the private loss with terms that depend on the public samples (e.g. a regularization term as in~\citep{golatkar2022mixed}, or a Bregman divergence term as in~\citep{amid2022public}). The second approach is to pre-train a model on public data, then fine-tune it on private data. This approach has proven successful in domains where public data is plentiful, as in image classification~\citep{de2022unlocking,mehta2023towards} and natural language~\citep{li2022large}. For a comprehensive survey, see~\citep{cummings2023challenges}. These works assume access to a separate public data set (often from the same or similar distribution as the private data). Our setting is different, in that we assume that the training data include some private and some public features. In this sense, our approach is orthogonal to, and can be combined with the aforementioned methods.
\paragraph{Label DP} Perhaps the closest setting to private training in the presence of public features is the notion of label DP~\citep{chaudhuri2011sample,ghazi2023regression,ghazi2021deep}. There, it is assumed that all features are public, and only labels are private. The setting of this paper bears some similarity with label DP, with two important distinctions: first, our setting aims to be more general, by allowing some but not all of the features to be public. Second, even when all features are public, our methods provide a stronger DP guarantee than label DP. This will be discussed in more detail in Section~\ref{sec:prelim}, but in short, our methods protect \emph{which row} of $\Xpub$ appears in a given training example (i.e., the correlation between the public features and the users), while label DP provides no such protection (it assumes that this is public knowledge).

\paragraph{Private model personalization} We will consider the class of multi-encoder models studied for example by~\cite{collins2021exploiting,singhal2021federated,jain2021private,shen2023share}. These models consist of learning a shared item encoder (that learns representations of items), together with a personalized user encoder (that learns individual preferences), as will be detailed in Section~\ref{sec:problem}. Prior work has only considered the case where all features are private. We study the case where the item features are public, while user features and labels are private, and show that by designing algorithms that adapt to this separation, one can achieve better utility. %They propose a family of algorithms based on alternating minimization, where one alternates between optimizing the shared encoder and the personalized classifier.  Our algorithms are also based on alternating minimization, but they differ in how the shared encoder is optimized. By taking advantage of the public-private feature separation, we achieve quality improvements both in theory and in practice.

Finally, in recent work,~\cite{curmei2023private} developed a method for private matrix completion with public features, a special case of our setting. They show strong empirical results (achieving the current SOTA on the DP MovieLens benchmarks), though their method is limited to matrix completion, and they do not provide a utility analysis. Our methods apply to a broader class of models, come with utility guarantees (albeit in a simplified case), and show substantial improvements on the same benchmarks.
\subsection{Notation}
Throughout the paper, $m,n$ will denote the number of items and users respectively. For a vector $x$, we will write $\|x\|_2$ for the Euclidean norm of $x$. For a matrix $X$, we will denote by $\|X\|_F$ its Frobenius norm, and by $\|X\|_2$ its induced norm. We will denote by $\Ncal^{d}$ a multivariate normal vector, and $\Ncal^{d\times d}$ a symmetric matrix whose upper triangle entries are i.i.d. normal. Additional notation is summarized in Appendix~\ref{app:notation}.
% ====================================================================
\section{Problem Formulation}
\label{sec:prelim}

\subsection{DP with Public Features}
First, we formalize the notion of private learning with public features. Let $\Xpub \in \Rbb^{m \times p}$ be a public feature matrix. To give a concrete example, one can think of rows of $\Xpub$ as representing items (for example movies to recommend, or ads to show to a user), and columns of $\Xpub$ as representing features, (for example, all possible movie genres and cast, or all possible ad features such as brand and price). Note that practical applications tend to have a large number of categorical features, and $\Xpub$ tends to be sparse.

We assume that each training example is of the form $(\xpriv, y, \xpub_j)$: it contains a private feature vector, a private label, together with some row $\xpub_j$ of $\Xpub$. For instance in a movie recommendation task, a training example might consist of a user's features $\xpriv$, the features of a movie that this user watched ($\xpub_j$), and a rating $y$ that the user assigned to the movie.

\begin{remark}
One important detail is that we want to protect \emph{which} row of the public matrix appears in each training example. In the above scenario, even though the features of a given movie are public, we want to protect which movies were seen by a user, as this is sensitive information. We formalize this by saying that $\Xpub$ is public, but the row index $j$ is private.
\end{remark}

Formally, we will define the private dataset as follows
\[
\Dcal = \{(\xpriv_i, y_i, j_i, k_i) \in \Rbb^{q} \times \Rbb \times  [m]\times[n]\}_{i \in \{1, \dots, \datasize\}}
\]
where $q$ is the dimension of private features, $m$ is the number of items, and $n$ is the number of users. Here $j_i$ represents an item index (it indexes into $\Xpub$), and $k_i$ represents a user index (its only purpose is to allow for user-level privacy accounting).

Given the dataset definition above, we adhere to the usual notion of (approximate) differential privacy:
\begin{definition}
A randomized algorithm $\Acal$ with output space $\Scal$ is said to be $(\epsilon, \delta)$-DP, if for all neighboring data sets $\Dcal, \Dcal'$, and all measurable $S \subset \Scal$,
\[
\Pr(\Acal(\Dcal) \in S) \leq e^\epsilon \Pr(\Acal(\Dcal') \in S) + \delta.
\]
\end{definition}
In example-level privacy, $\Dcal, \Dcal'$ are said to be neighboring data sets if one is obtained from the other by removing a single example. In user-level privacy, they are said to be neighboring if one is obtained from the other by removing all examples belonging to a user.

\begin{remark}[Interpretation of DP with Public Features] Note that we use the standard DP definition. The only new assumption we make is that the private data includes an index $j_i \in [m]$, used to lookup a feature vector from a public matrix $\Xpub$. The fact that $\Xpub$ is public means in particular that it doesn't change with the dataset $\Dcal$.
\end{remark}

\begin{remark}[Comparison to Label DP]
Similar to our setting, label DP~\citep{chaudhuri2011sample,ghazi2021deep} also defines a privacy guarantee where only part of the data needs protection. There are two distinctions with our setting: first, label DP assumes all features to be public, while DP with public features (our setting) allows for some features to be private. Second, even in the special case where all features are public, %TOOD: add that labels are private.
our setting provides a stronger notion of privacy. This is due to the definition of neighboring data sets: in label DP, neighboring data sets are only allowed to differ in a single label, and feature vectors are not allowed to change. As a consequence, participation in the process is not protected. In our setting, participation is protected (since completely removing the example yields a valid neighboring dataset). Another consequence is that label DP provides no protection of \emph{which} public item appears in an example (again, since the feature vector is not allowed to change), while our setting does (the index $j_i$ can change).
\end{remark}

% ====================================================================
\subsection{Multi-Encoder Models and Alternating Minimization}
\label{sec:problem}
We consider multi-encoder models, studied for example by~\cite{agarwal2009regression,jain2021private,shen2023share}, and commonly used in practice, in advertising~\citep{agarwal2009regression}, recommender systems~\citep{covington2016deep,volkovs2017dropoutnet}, and similarity learning~\citep{chopra2005learning,schroff2015facenet,dong2018triplet}.

In our case, given a training example $(\xpriv, \xpub)$, the model's prediction is given by
\begin{equation}
\label{eq:multi-encoder}
f_{\theta_u, \theta_v}(\xpriv, \xpub) = u_{\theta_u}(\xpriv) \cdot v_{\theta_v}(\xpub),
\end{equation}
where $v_{\theta_v} : \Rbb^{p} \to \Rbb^{d}$ is an encoder that maps the public features to an embedding in $\Rbb^d$, $u_{\theta_u} : \Rbb^{q} \to \Rbb^{d}$ is a second encoder that maps the private features to another embedding in $\Rbb^d$, and the model's prediction is the dot product of the two embeddings. Finally, $\theta_v \in \Rbb^{d_v}, \theta_u\in \Rbb^{d_u}$ are the parameters of the two encoders. We will refer to $v_{\theta_v}(\cdot)$ as the public encoder (since it operates on public features) and to $u_{\theta_u}(\cdot)$ as the private encoder. This architecture is well-suited to our problem, as it creates a separation between the private and public features, and as we shall see, our algorithms will take advantage of this separation.

\begin{remark}[Personalized model interpretation] One interpretation of this model (see~\cite{jain2021private,shen2023share}), is that the prediction task is decomposed into training a shared \emph{item encoder} ($v_{\theta_v}$) which learns a user-independent representation of items, together with a \emph{personalized classifier} ($u_{\theta_u}$) which captures each user's preferences. This point of view is also common in the collaborative filtering literature~\citep{hu2008ials}.
\end{remark}

The goal is to minimize the empirical risk
\begin{equation}
\label{eq:loss}
\sum_{i = 1}^D \ell\left(u_{\theta_u}(\xpriv_i) \cdot v_{\theta_v}(\xpub_{j_i}), y_i\right),
\end{equation}
where $\ell$ is a loss function. A popular approach to solve this problem is the Alternating Minimization (AM) procedure described in Algorithm~\ref{alg:am}, where one alternates between optimizing one encoder (while the other is frozen) and vice-versa. The recent works of~\cite{chien2021private,jain2021private,shen2023share,curmei2023private} all fall in this family of algorithms, and they only differ in how the encoders are optimized (i.e. in the sub-routines \userUpdate{} and \itemUpdate). %The interested reader can refer to the appendix for a comparison of DP alternating minimization methods.
None of them, however, considers the problem of training with public features. Our focus will be to design an algorithm for optimizing the \emph{public encoder}. For the private encoder, any DP training procedure such as DP-SGD can be used. % (or, in the special case where the user encoder consists of an independent embedding per user, one can use non-private minimization), 

\begin{algorithm}[t]
\caption{Alternating Minimization}
\label{alg:am}
\SetAlgoLined
\DontPrintSemicolon
\SetKwProg{proc}{Procedure}{}{}
{\bf Inputs}: Training data $\Dcal = \{(\xpriv_i, y_i, j_i, k_i)\}_{i \in \{1, \dots, \datasize\}}$, number of outer steps $S^{\text{outer}}$\\
\For{$1\leq s\leq S^{\text{outer}}$}{
$\theta_u \leftarrow \userUpdate(\theta_u, \theta_v, \Dcal)$\\
$\theta_v \leftarrow \itemUpdate(\theta_u, \theta_v, \Dcal)$
}
\KwRet $\theta_u, \theta_v$
\end{algorithm}

% \begin{remark}
% Alternating minimization has also been studied in the Federated Learning (FL) literature~\citep{singhal2021federated,collins2021exploiting}, where the goal is to perform distributed training while keeping each user's data on the user's device. Our presentation will focus on the centralized setting, but we will discuss, in the appendix, implications of our algorithms in the FL setting.
% \end{remark}

% ===========================================================
\section{Private Training of the Public Encoder}
\label{sec:training}
In this section, we design and analyze algorithms for optimizing the public encoder under a quadratic loss, i.e., $\ell(\hat y, y) = \frac{1}{2}(\hat y - y)^2$. We discuss in Appendix~\ref{app:convex} the more general case where $\ell$ is convex.

For ease of notation, we will denote by $u_i = u_{\theta_u}(\xpriv_i)$. We can then write the loss~\eqref{eq:loss} in a more concise form:
\begin{equation}
\label{eq:loss_quad}
\Lcal(\theta_v) = \frac{1}{2}\sum_{i = 1}^{\datasize} \left(u_i \cdot v_{\theta_v}(\xpub_{j_i}) - y_i\right)^2.
\end{equation}
Note that since the private encoder is frozen, we only keep the explicit dependence on the public encoder parameters $\theta_v$.

% -------------------------------------------------------------
\subsection{Gradient Decomposition under Public Features}
The main observation is that we can factorize the gradient into terms that only depend on public features, and terms that only depend on private data. Let $\Oj = \{i \in [D] : j_i = j\}$ (in other words, $\Oj$ is the set of examples which contain item $j$). Then we have the following decomposition:
\begin{proposition}
\label{prop:decomposition}
The gradient of the loss~\eqref{eq:loss_quad} is equal to:
\begin{equation}
\label{eq:grad}
\nabla \Lcal(\theta_v) = \sum_{j = 1}^m \frac{\partial v_{\theta_v}(\xpub_j)}{\partial \theta_v}[A_j v_{\theta_v}(\xpub_j) - b_j],
\end{equation}
where, for all $j$,
\begin{equation}
\label{eq:Aj_bj}
A_j = \sum_{i \in \Oj} u_i u_i^\top, \quad\quad b_j = \sum_{i \in \Oj} y_i u_i.
\end{equation}
\end{proposition}
Observe that in~\eqref{eq:grad}, both of the terms $v_{\theta_v}(x_j) \in \Rbb^d$ and $\partial v_{\theta_v}(x_j)/\partial \theta_v \in \Rbb^{d_v \times d}$ only depend on public features (these terms correspond respectively to the public encoder's forward pass, and its Jacobian), while the terms $A_j \in \Rbb^{d \times d}, b_j \in \Rbb^d$ depend on private data ($y_i$ is a private label, and $u_i = u_{\theta_u}(\xpriv_i)$ is a function of the private features). We will refer to $A_j, b_j$ as the sufficient statistics for item $j$. $A_j$ can be viewed as a partial private covariance matrix (partial in the sense that it's the covariance of just the examples that involve the $j$-th row of the public matrix).

% -------------------------------------------------------------
\subsection{Sufficient Statistics Perturbation Algorithms}
\label{sec:alg}
The gradient decomposition in Proposition~\ref{prop:decomposition} suggests that one can achieve privacy protection by adding noise to the terms $A_j, b_j$, instead of the final gradient. We propose two algorithms based on this approach, see Algorithms~\ref{alg:ssp} and~\ref{alg:ssp-corr}.

Both operate by first computing the sufficient statistics for all items, then taking $T$ steps of gradient descent, where the gradient is computed based on the noised statistics. They differ in how the noise is added (the differences are highlighted in blue). The first algorithm samples independent noise at each iteration, while the second only adds noise once, and reuses the noisy statistics across all iterations. Notice that to achieve the same privacy, the noise standard deviations $\sigma$ needed in the two algorithms are different.
% -------------------------------------------------------------
\begin{algorithm}[h]
%TODO: rename to SSP-cor and SSP-ind.
\caption{SSP1: Sufficient Statistics Perturbation with Independent Noise}
\label{alg:ssp}
\SetAlgoLined
\DontPrintSemicolon
\SetKwProg{proc}{Procedure}{}{}
{\bf Inputs}: Public features $\Xpub$, training data $\Dcal = \{(\xpriv_i, y_i, j_i, k_i)\}_{i \in \{1, \dots, \datasize\}}$, optional weights\footnotemark $\{w_i\}$, number of steps $T$, clipping parameters $\Gamma_y, \Gamma_u$, noise standard deviation $\sigma$, learning rate $\eta_t$, initial parameters $\theta_v^{0}, \theta_u^0$.\\
Let $\bar u_i = \clip(u_{\theta_u}(\xpriv_i), \Gamma_u)$,
$\bar y_{i} = \clip(y_{i}, \Gamma_y)$\\
%{\bf Compute sufficient statistics:}\\
\For{$1 \leq j \leq m$}{
$A_j \leftarrow \sum_{i \in \Oj} w_i\bar u_i \bar u_i^\top$\\
$b_j \leftarrow \sum_{i \in \Oj} w_i\bar y_i \bar u_i$
}
\For{$0 \leq t \leq T-1$}{
\blue{
\For{$1 \leq j \leq m$\label{line:ssp1_noise_begin}}{
$\hat A^t_j = A_j + \sigma\Gamma_u^2 \Ncal^{d \times d}$\\
$\hat b^t_j = b_j + \sigma\Gamma_y\Gamma_u \Ncal^{d}$\label{line:ssp1_noise_end}\\
}
}
$\hat G^t \leftarrow \sum_{j = 1}^m \frac{\partial v_{\theta_v^t}(\xpub_j)}{\partial \theta_v}[\hat A_j^t v_{\theta_v^t}(\xpub_j) - \hat b_j^t]$\label{line:ssp1_grad}\\
$\theta_v^{t+1} \leftarrow \theta_v^t - \eta_t \hat G^t$
}
\KwRet $\theta_v^T$
\end{algorithm}

% -------------------------------------------------------------

\begin{algorithm}[h!]
\caption{SSP2: Sufficient Statistics Perturbation with Correlated Noise}
\label{alg:ssp-corr}
\SetAlgoLined
\DontPrintSemicolon
\SetKwProg{proc}{Procedure}{}{}
{\bf Inputs}: Public features $\Xpub$, training data $\Dcal = \{(\xpriv_i, y_i, j_i, k_i)\}_{i \in \{1, \dots, \datasize\}}$, optional weights $\{w_i\}$, number of steps $T$, clipping parameters $\Gamma_y, \Gamma_u$, noise standard deviation $\sigma$, learning rate $\eta_t$, initial parameters $\theta_v^{0}, \theta_u^0$.\\
Let $\bar u_i = \clip(u_{\theta_u}(\xpriv_i), \Gamma_u)$,
$\bar y_{i} = \clip(y_{i}, \Gamma_y)$\\
%{\bf Compute sufficient statistics:}\\
\For{$1 \leq j \leq m$}{
$\hat A_j \leftarrow \sum_{i \in \Oj} w_i\bar u_i \bar u_i^\top \blue{+ \sigma\Gamma_u^2 \Ncal^{d \times d}}$\label{line:Aj}\\
$\hat b_j \leftarrow \sum_{i \in \Oj} w_i\bar y_i \bar u_i \blue{+ \sigma\Gamma_y\Gamma_u \Ncal^d}$\label{line:bj}.
}
%{\bf Optimize the item encoder:}\\
\For{$0 \leq t \leq T-1$}{
$\hat G^t \leftarrow \sum_{j = 1}^m \frac{\partial v_{\theta_v^t}(\xpub_j)}{\partial \theta_v}[\hat A_j v_{\theta_v^t}(\xpub_j) - \hat b_j]$ \label{line:ssp2_grad}\\
$\theta_v^{t+1} \leftarrow \theta_v^t - \eta_t \hat G^t$
}
\KwRet $\theta_v^T$
\end{algorithm}
% -------------------------------------------------------------

\begin{remark}\label{rem:reusing_noise}Reusing the noisy statistics is only possible due to the fact that $A_j, b_j$ \emph{do not depend on the iterate $\theta_v^t$} (see eq.\eqref{eq:Aj_bj}), so they don't need to be recomputed. One consequence is that the scale of the noise in Algorithm~\ref{alg:ssp-corr} is independent of the number of steps $T$ (as will be apparent in the privacy guarantee). However, this makes the noise in the gradient estimates correlated across iterations, which makes 
utility analysis difficult. For this reason, our utility analysis will be for Algorithm~\ref{alg:ssp} (the independent noise version). Empirically we evaluate both algorithms, and we find that the correlated version (Algorithm~\ref{alg:ssp-corr}) performs better.
\end{remark}

To give some intuition how this approach improves upon noisy GD, observe that the gradient is a vector in $\Rbb^{d_v}$ (where $d_v$ is the number of parameters of the encoder, which can be very large), while the right factor in the decomposition~\eqref{eq:grad} (the term $A_jv_{\theta_v}(\xpub_j) + b_j$) is a vector in $\Rbb^{d}$, where $d$ is the \emph{output dimension} of the encoder, and typically much smaller than $d_v$. Protecting the lower dimensional object is more efficient, as will become clear in the analysis.

\footnotetext{The optional weights $w_i$ (used as weights in the sufficient statistics, see Lines~\ref{line:Aj}-\ref{line:bj}) are useful in practice for user-level privacy: for example, by reducing the weights of a user who has many examples, this allows us to control the worst-case user sensitivity. See Theorem~\ref{thm:priv_user} for a precise statement on how weights affect the DP guarantee.}

\paragraph{Preserving Gradient Sparsity} Another practical advantage of adding noise to the sufficient statistics is that it allows us to preserve gradient sparsity. Sparse gradients are very common in models that use a large number of categorical features, which is the case in most personalization models~\citep{cheng2016wide}. Typically, only few features are active for a given item (and inactive features have 0 gradients). This sparsity will manifest in the Jacobian term in eq.\eqref{eq:grad}, indeed, if the encoder is of the form $v(\xpub_j) = \nu(\theta_{in}^\top\xpub_j)$ ($\theta_{in}$ represents an input embedding layer, and $\nu$ is the rest of the encoder), then its Jacobian w.r.t. $\theta_{in}$ is of the form $\xpub_j \rho_j^\top$ for some vector $\rho_j$, meaning that for any feature that is not active ($\xpub_{jl} = 0$) the corresponding $l$-th row in the Jacobian (and hence in the gradient) is also $0$. See Appendix~\ref{app:exp} for an illustration of the sparsity in our experiments.
% CAMERA-READY VERSION replace last sentence by: See Figure~\ref{fig:feature_density} in Appendix~\ref{app:datasets} for an illustration of the sparsity in our experiments.

In noisy gradient descent and its variants, since noise is added to the final gradient, this destroys its sparsity--this was identified as one of the challenges of applying DP-SGD in practice~\citep{zhang2021wide}. In our SSP algorithms, since no noise is added to the Jacobian, its sparsity (and that of the gradient) are preserved.

% -------------------------------------------------------------
\subsection{Privacy Guarantees}
We now state the privacy guarantee for both algorithms. All proofs are deferred to the appendix.%The guarantees will be stated in terms of zero-concentrated DP (zCDP), translation from zCDP to $(\epsilon, \delta)$-DP is standard~\citep{bun2016concentrated}[Proposition 1.3].
\begin{theorem}[Example-level Privacy Guarantee]
\label{thm:priv_example}
Let $w_i = 1$ for all $i$. Let $\epsilon, \delta > 0$ with $\epsilon < \log 1/\delta$. Then Algorithm~\ref{alg:ssp} (resp. Algorithm~\ref{alg:ssp-corr}) with standard deviation $\sigma = \frac{\sqrt{8T\log 1/\delta}}{\epsilon}$ (resp. $\sigma = \frac{\sqrt{8\log 1/\delta}}{\epsilon}$) is $(\epsilon, \delta)$-DP.
\end{theorem}
The main difference between the two algorithms is that noise scales with $\sqrt{T}$ in the independent noise version.

To state the user-level guarantee, it will be useful to define the set $\Ok = \{i : k_i = k\}$ (in other words, these are the indices of examples that belong to user $k$).
\begin{theorem}[User-level Privacy Guarantee]
\label{thm:priv_user}
Let $\bar w^2 = \max_{k} \sum_{i \in \Omega^k} w_i^2$. Let $\epsilon, \delta > 0$ with $\epsilon < \log 1/\delta$. Then Algorithm~\ref{alg:ssp} (resp. Algorithm~\ref{alg:ssp-corr}) with noise standard deviation $\sigma = \frac{\bar w\sqrt{8T\log 1/\delta}}{\epsilon}$ (resp. $\sigma = \frac{\bar w\sqrt{8\log 1/\delta}}{\epsilon}$) is $(\epsilon, \delta)$-DP.
\end{theorem}

% -------------------------------------------------------------
\subsection{Utility Analysis for Linear Encoders}
Our goal in this section is to identify settings in which access to public features improves utility compared to the same problem where all features are private. We will consider linear encoders, i.e. $v_{\theta_v}(\xpub_j) = \theta_v^\top \xpub_j$ where $\theta_v \in \Rbb^{p \times d}$. The problem is to minimize
$
\Lcal(\theta_v) = \sum_{i = 1}^\datasize ({\xpub_{j_i}}^\top \theta_v u_i - y_i)^2
$,
and the gradient factorization in Proposition~\ref{prop:decomposition} simplifies to
\[
\nabla \Lcal(\theta_v) = \sum_{j = 1}^m \xpub_j[A_j \theta_v^\top\xpub_j - b_j]^\top \in \Rbb^{p \times d}.
\]
\begin{assump}
\label{assump:1}
In this section, we assume that there exist $\Gamma_x, \Gamma_y, \Gamma_u$ such that for all~$j$, $\|\xpub_j\|_2 \leq \Gamma_x$, and for all $i$, $\|u_i\|_2 \leq \Gamma_u$ and $|y_i| \leq \Gamma_y$. Furthermore, we will optimize the problem over a bounded set $\Theta = \{\theta \in \Rbb^{p \times d} : \max_j \|\theta^\top \xpub_j\|_2 \leq \Gamma_y/\Gamma_u\}$.
\end{assump}

The boundedness assumptions on $y,u$ are standard in private linear regression~\citep{wang2018revisiting}. The definition of the feasible set $\Theta$ simply requires that the model's output ($\xpub_j \theta^\top u_i$) be of the same scale as the labels--this is often enforced in practice by normalizing the output of the encoder. This condition is for convenience (so that certain constants simplify).

We have the following bound on excess empirical risk:
\begin{theorem}[Utility Guarantee of Algorithm~\ref{thm:utility}]
\label{thm:utility}
Suppose Assumption~\ref{assump:1} holds. Let $\Gamma = \Gamma_x\Gamma_y\Gamma_u$. Let $\theta_v^* = \argmin_{\theta_v \in \Theta} \Lcal(\theta_v)$, and let $\hat \theta_v$ be the output of projected SSP1 run with $\sigma = \rho\sqrt{T}$, for $T = \frac{D^2}{md\rho^2}$ steps and with learning rates $\eta_t = \frac{|\Theta|}{\Gamma\datasize \sqrt {8t}}$. Then
\[
\Exp[\Lcal(\hat \theta_v)] - \Lcal(\theta_v^*) = \tilde\Ocal\left(|\Theta|\Gamma \rho \sqrt{md}\right),
\]
where $\tilde O$ hides poly-log factors. Furthermore, setting $\rho = \frac{\sqrt{8\log 1/\delta}}{\epsilon}$ guarantees that the algorithm is $(\epsilon, \delta)$-DP.
\end{theorem}

To highlight the implications of Theorem~\ref{thm:utility}, we compare the bound to (projected) noisy gradient descent~\citep{bassily2014erm}, where at each iteration
\[
\theta_v^{t+1} = \proj_{\Theta}\left[\theta_v^t - \eta \sum_{i = 1}^\datasize \clip(g_i(\theta_v^t), \Gamma) + \sigma\Gamma\Ncal^{p \times d}\right],
\]
where $g_i(\theta_v) = \xpub_{j_i}[u_iu_i^\top \theta_v^\top\xpub_j - y_iu_i]^\top$ (note that under Assumption~\ref{assump:1}, we have the bound $\|g_i\|_2 \leq \Gamma = \Gamma_x\Gamma_u\Gamma_y$, hence we use $\Gamma$ as the clipping constant). A standard analysis~\citep{bassily2014erm} shows that the excess empirical risk for noisy GD is bounded by $\tilde\Ocal(|\Theta|\Gamma\rho\sqrt{pd})$. The main difference in Theorem~\ref{thm:utility} is that the dimension dependence $\sqrt{pd}$ is replaced with $\sqrt{md}$. Recall that $d$ is the output dimension of the encoder, $m$ is the number of public items, and $p$ is the dimension of public features. For data with many categorical features, $p$ can be orders of magnitude larger than $m$, especially when one uses feature crosses. In such cases, the bound of Theorem~\ref{thm:utility} suggests potentially large utility improvements.

\subsection{Other Practical Considerations}
\label{sec:complexity}
\paragraph{Mini-batching} Our algorithms were described in the full-batch setting for simplicity. In this section, we discuss their mini-batched variants.

In SSP1, this can be achieved by replacing Lines~\ref{line:ssp1_noise_begin}-\ref{line:ssp1_noise_end} with the following: uniformly sample a batch of examples $B \subset [\datasize]$, then compute the sufficient statistics over the batch, see Algorithm~\ref{alg:mini-ssp} in the appendix.%i.e., $\hat A_j = \sum_{i \in \Oj \cap B} u_iu_i^\top + \sigma \Gamma_u^2 \Ncal^{d\times d}$, and $\hat b_j = \sum_{i \in \Oj \cap B} y_iu_i + \Gamma_y\Gamma_u \Ncal^{d}$.

In SSP2, mini-batching can be achieved by replacing Line~\ref{line:ssp2_grad} with the following: uniformly sample a batch $B \subset [m]$ of \emph{items}, then let $\hat G^t = \sum\limits_{j \in B} \frac{\partial v_{\theta_v^t}(\xpub_j)}{\partial \theta_v}(\hat A_j v_{\theta_v^t}(\xpub_j) - \hat b_j)$, see Algorithm~\ref{alg:mini-ssp-corr} in the appendix. Note that in this case, we sample items instead of training examples, which in many cases has a computational advantage that we discuss below.

This per-item sampling is not recommended, however, in DP-SGD or SSP1, because sampling per-item precludes amplification by sampling~\citep{bassily2014erm, abadi2016dpsgd} which would lead to worse privacy guarantees. Per-item sampling is possible in SSP2 because we only add noise once (so there is no need for amplification).

%The mini-batched variants of SSP1, SSP2, and DP-SGD are stated in Algorithms~\ref{alg:mini-ssp},~\ref{alg:mini-ssp-corr},~\ref{alg:dpsgd}.
%We emphasize that we sample \emph{items} instead of the typical sampling of training records, since the gradient decomposition (Proposition~\ref{prop:decomposition}) is a sum over items. In a sense, the decomposition relies on a partitioning of the training data by public item, $[D] = \sqcup_{j = 1}^m \Oj$: the $j$-th term represents the total contribution of all records that involve item $j$.

%However, this per-item sampling is challenging in SSP1: since we re-apply noise at every iteration, one needs to use amplification by sampling to obtain tighter privacy guarantees~\citep{bassily2014erm, abadi2016dpsgd}, but existing amplification techniques assume per-record sampling, and don't apply to per-item sampling (this breaks independence between records). SSP2, however, circumvents the issue entirely: since each statistic is noised only once, it can be used across mini-batches without paying any additional privacy cost--amplification is irrelevant in this case.

\paragraph{Computational Complexity} 
Per-item sampling in SSP2 has important implications on computational cost, which we now discuss.

First, notice that the gradient of an example $i$ is $\Jcal_{j_i}[u_iu_i^\top v_{j_i} - y_iu_i]$ (where we write $v_j = v_{\theta_v}(\xpub_j)$ for the encoder's output, and $\Jcal_j := \frac{\partial v_{\theta_v}(\xpub_j)}{\partial \theta_v}$ for the Jacobian of item $j$). This involves computing a forward pass $v_j$ and a partial back-propagation $\Jcal_j$. Under per-record sampling (DP-SGD or SSP1), an item $j$ is revisited many times during one epoch (since it appears in all records in $\Oj$), so the forward/backward passes for this item are recomputed at each visit. Whereas under per-item sampling (SSP2), the forward/backward passes are computed exactly once per item per epoch, reducing the cost of gradient computation. At the same time, SSP has the additional overhead of computing the sufficient statistics, but since this is done once, one can hope to amortize it across iterations.

The following proposition compares the computational cost of SSP2 and DP-SGD--more precisely, we will compare to an optimized version of DP-SGD (that avoids recomputing the forward/backward passes if the same item appears more than once in the batch, see Algorithm~\ref{alg:dpsgd} in the appendix).

\begin{proposition}
\label{prop:complexity}
Let $c$ be the average cost of computing one forward/backward pass of the public encoder. Let $\beta_j$ be the expected number of batches in which item $j$ appears per epoch. Then the total expected cost over $e$ epochs is bounded as follows:
The cost of DP-SGD (Algorithm~\ref{alg:dpsgd}) is $\Omega(ce \sum_{j=1}^m \beta_j)$, and the cost of SSP2 (Algorithm~\ref{alg:mini-ssp-corr}) is $\Ocal(\datasize d^2 + (c+d^2)em)$.
\end{proposition}

We give a detailed comparison in Appendix~\ref{app:complexity} under different regimes. Here, we discuss a simplified case. Suppose that $e \geq \datasize/m$, then the cost of SSP2 becomes $\Ocal((c+d^2)em)$, and we can more easily compare it to DP-SGD. The ratio (cost of SSP2 by cost of DP-SGD) is $\Ocal\left(\frac{m(c+d^2)}{c \sum_{j = 1}^m \beta_j}\right) = \Ocal(\frac{m}{\beta}(1 + \frac{d^2}{c}))$, where $\beta = \sum_{j = 1}^m \beta_j$. The first term $\frac{m}{\beta}$ is smaller than $1$ (typically orders of magnitude smaller): indeed, the quantity $\beta$ is at least $m$ (the full batch case) and at most $\datasize$ (when the batch size is $1$). It can be estimated precisely if one knows the item counts (we show examples in Appendix~\ref{app:complexity}).

Finally, to bound the term $\frac{d^2}{c}$, notice that if the encoder has at least one hidden layer of width $\geq d$ (a reasonable assumption, since $d$ is the encoder's output dimension), then $c \geq d^2$, and the ratio simplifies to $\Ocal(\frac{m}{\beta})$ which, as argued above, can be very small. Our experiments fall under this setting, and the improvement in our case is approximately two orders of magnitude (see next section).

%Even for a simple linear encoder without hidden layers, one has $c \geq sd$ where $s$ is the number of non-zero entries in the feature vector. When $s = 1$ (a single active feature per item), SSP2 is expected to be slower than DP-SGD. If $s$ is comparable to $d$, SSP2 is expected to be faster.

% -------------------------------------------------------------
\section{Experimental Results}
\label{sec:exp}
We evaluate our methods on two standard private recommendation benchmarks~\citep{jain2018differentially,chien2021private,krichene2023multi,curmei2023private}, based on the MovieLens data sets~\citep{harper16movielens}. The first is a regression task on MovieLens 10M (abbreviated as ML10M in the results), and the second is a classification task on MovieLens 20M (abbreviated as ML20M).

% -------------------------------------------------------------
% -------------------------------------------------------------
% -------------------------------------------------------------
\begin{figure*}[h!]
\centering
\setkeys{Gin}{width=\linewidth}
\begin{tabularx}{\linewidth}{XX}
\includegraphics{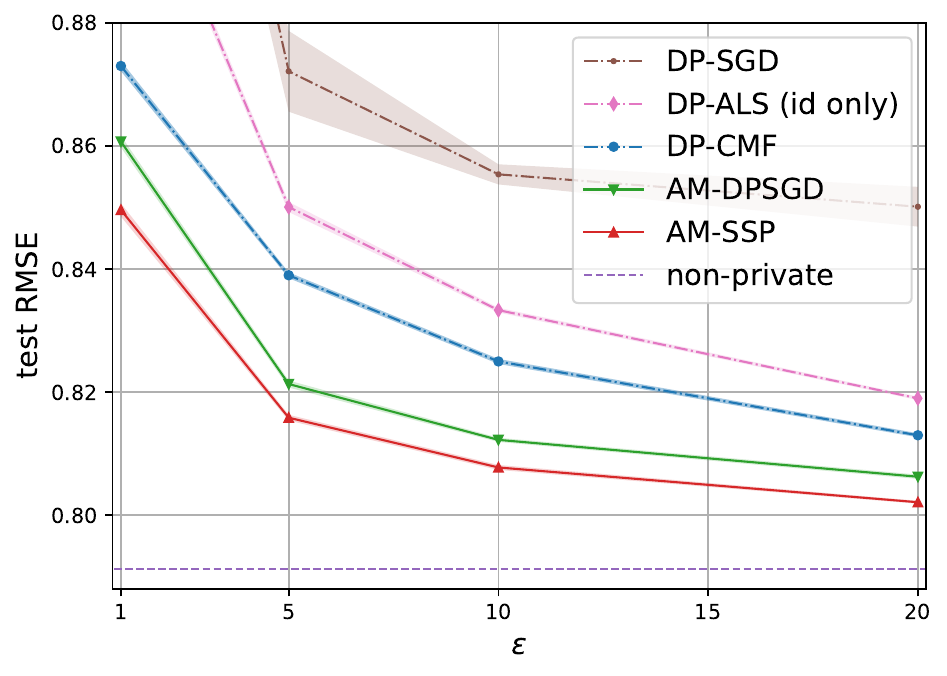}
\caption{Privacy-RMSE trade-off in ML10M (lower is better).}
\label{fig:ml10m}
    &
\includegraphics[]{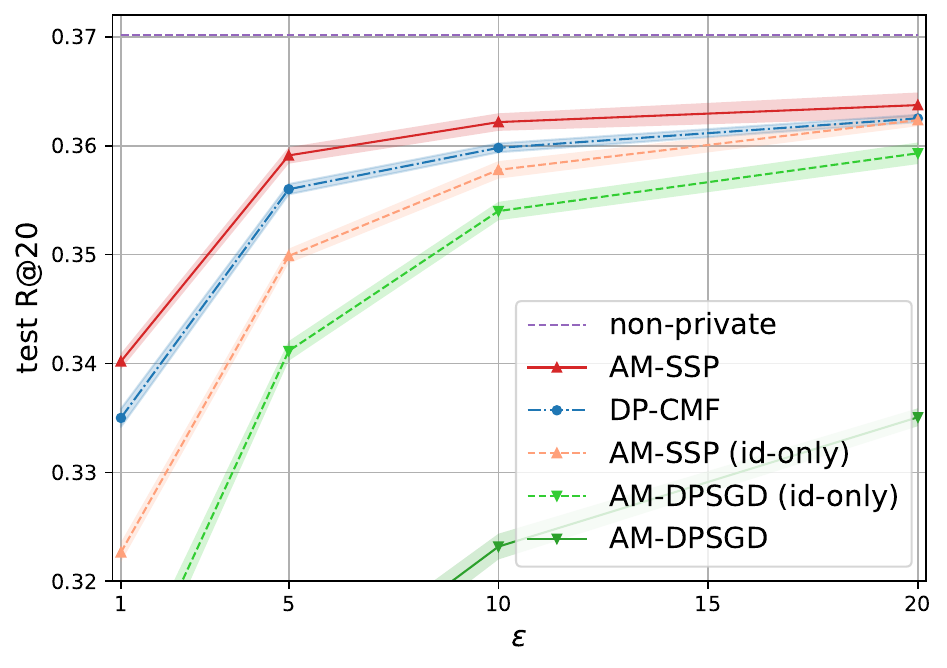}
\caption{Privacy-Recall trade-off in ML20M (higher is better).}
\label{fig:ml20m}
\end{tabularx}
\end{figure*}
% -------------------------------------------------------------
% -------------------------------------------------------------
% -------------------------------------------------------------

\subsection{Experimental Setup}
\paragraph{Data} Each training example is a tuple $(\xpriv_i, \xpub_{j_i}, y_i)$ where $\xpriv_i$ is a user's id, $\xpub_{j_i}$ is the feature vector of a movie, and $y_i$ is a rating that the user assigned to the movie. The public features $\Xpub$ consist of metadata contained in the original MovieLens data (movie genre and release year) along with a richer set of metadata extracted from Wikidata by~\cite{curmei2023private}, containing information such as movie cast and fine-grained genres. Statistics about the data and features are detailed in Appendix~\ref{app:datasets}.

\paragraph{Models and Algorithms} The current SOTA on these DP benchmarks consist of matrix completion methods: the DP-ALS algorithm of~\cite{chien2021private} (which does not use features, only movie ids), and the DP-CMF algorithm of~\cite{curmei2023private} (which factorizes a concatenation of the rating matrix and the public feature matrix). Matrix completion is a special case of the multi-encoder setting~\eqref{eq:multi-encoder}, where both encoders are linear and the feature vectors are one-hot.

In our experiments, we keep the same id-based user encoder, but train a more general item encoder, consisting of a wide embedding layer $\in \Rbb^{p \times d}$ (which embeds the categorical features) followed by a dense layer $\in \Rbb^{d \times d}$. The model is trained using alternating minimization (AM, see Algorithm~\ref{alg:am}) where \userUpdate{} is the same as in DP-ALS/DP-CMF, and we evaluate different methods for the \itemUpdate{}\footnote{Since we design algorithms for training the public encoder, our comparison uses the same user encoder and the same \userUpdate{}, so quality differences can be attributed to the public encoder and the \itemUpdate{}.}. We consider two variants:
AM-DPSGD uses DP-SGD for the \itemUpdate{}, and AM-SSP uses SSP2 (Algorithm~\ref{alg:ssp-corr}) for the \itemUpdate{} (we also compare the SSP1 and SSP2 variants).

\paragraph{Evaluation Protocol} We follow the exact protocol used in prior work on these benchmarks~\citep{jain2018differentially,chien2021private,krichene2023multi,curmei2023private} and use their code for data processing and privacy accounting. Each data set is split into training/validation/test, we tune hyper-parameters on the validation data and report metrics on test data (the shaded region on each plot shows the standard deviation across ten runs). Utility is measured using RMSE on ML10M and Recall@20 on ML20M. We report \emph{user-level} $(\epsilon, \delta)$ privacy guarantees, where $\delta = 10^{-5}$ for ML10M and $8.10^{-6}$ for ML20M. A fully detailed account of the setup (and additional experiments) can be found in Appendix~\ref{app:exp}.

% -------------------------------------------------------------
\subsection{Privacy Utility Trade-off}
We compare the privacy/utility trade-offs of the different algorithms. The results for ML10M are reported in Figure~\ref{fig:ml10m}. We start by observing that feature-based methods (DP-CMF, AM-DPSGD and AM-SSP) all improve upon the DP-ALS baseline (which uses only item ids). Our method (AM-SSP) significantly improves upon the prior SOTA (DP-CMF), bringing utility much closer to the non-private baseline. The absolute improvement ranges from 0.025 (at $\epsilon = 1$) to 0.012\footnote{To put this into perspective, an absolute improvement of 0.004 on the ML10M benchmark is considered significant, and some works have reported even smaller improvements, see~\citep{rendle2019baselines} for a survey.} (at $\epsilon = 20$).

Turning to the ML20M benchmark (Figure~\ref{fig:ml20m}), we see a more modest (though non-trivial) improvement compared to DP-CMF. One interesting observation, however, is that AM-DPSGD performs quite poorly on this benchmark: observe the large gap between AM-SSP and AM-DPSGD. To further investigate this performance gap, we apply the same methods to an id-only model that doesn't use any of the item features (the results are in dashed lines on the same figure). Remarkably, adding features improves quality under AM-SSP, but it \emph{degrades it under AM-DPSGD}. Recall that SSP takes explicit advantage of the fact that movie features are public, while DPSGD doesn't. This is an example where adapting to the public-private feature separation leads to substantial gains.

% -------------------------------------------------------------
\subsection{Computational Cost}
\begin{figure*}[t]
\centering
\setkeys{Gin}{width=\linewidth}
\begin{tabularx}{\linewidth}{XX}
\includegraphics[]{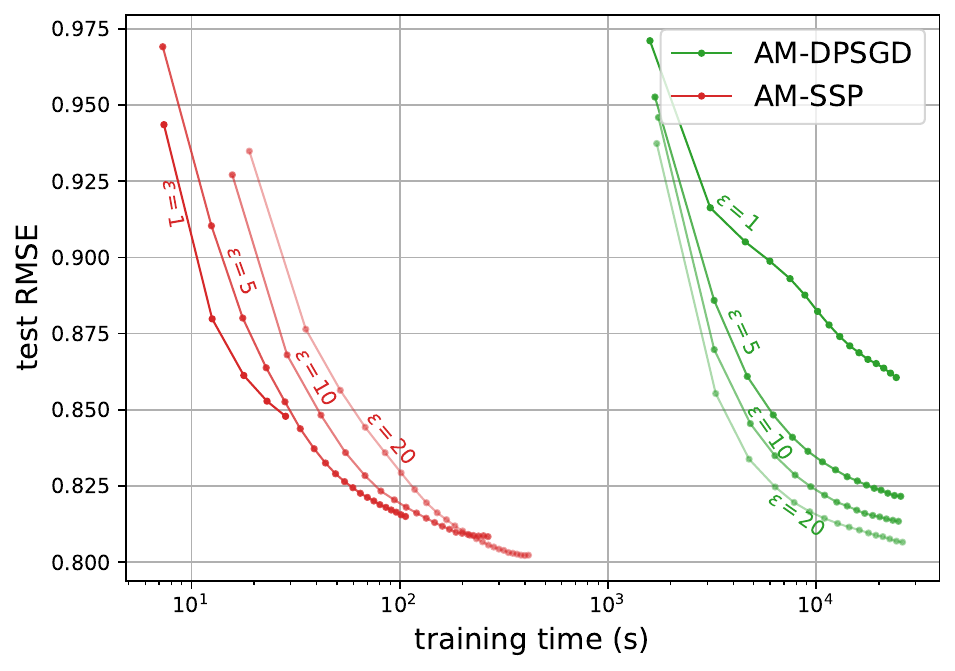}
\caption{Time-evolution of test RMSE on ML10M.}
\label{fig:traj}
    &
\includegraphics[]{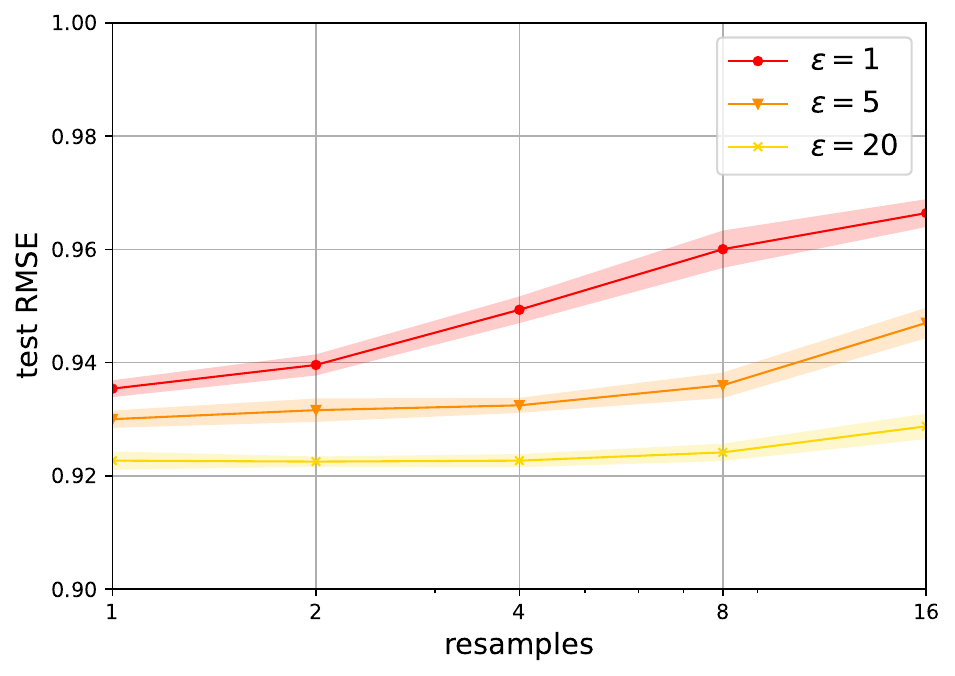}
\caption{Comparison of SSP1 and SSP2 on ML10M.}
\label{fig:comp}
\end{tabularx}
\end{figure*}

Besides quality improvements, we also highlight the computational advantage of AM-SSP. We plot, in Figure~\ref{fig:traj}, the time-evolution of the test RMSE for AM-SSP and AM-DPSGD (with the optimal hyper-parameters tuned separately for each method). The results show approximately two-orders of magnitude difference in training time between the two methods. This illustrates the complexity advantage of SSP discussed in Section~\ref{sec:complexity}. The extent of the difference will generally depend on the problem parameters (such as number of items, sparsity, and batch size). This example shows that the improvement can be quite significant in practice (similar results are shown in the appendix for ML20M).
% -------------------------------------------------------------
% -------------------------------------------------------------
\subsection{Comparing SSP1 and SSP2}
Finally, we assess the empirical difference between the SSP1 and SSP2 variants (Algorithms~\ref{alg:ssp}-\ref{alg:ssp-corr}), see Figure~\ref{fig:comp}. In this experiment, we only optimize the public encoder (i.e., without alternating minimization). Recall that SSP1 resamples noise at every step, while SSP2 adds noise once and reuses the statistics across iterations. Here we experiment with varying the number of times noise is resampled, so resamples=1 corresponds to SSP2, while resamples=16 (16 is the total number of steps) corresponds to SSP1. We find that as we increase the resampling frequency, quality degrades, and this appears to be more significant for lower $\epsilon$ (similar results hold on ML20M).

% -------------------------------------------------------------
\section{Discussion}

% \paragraph{Separating Sensitive and Public Features} In certain applications, it may not be possible to fully separate user and item features, specifically in the presence of a feature cross between a private and public feature. One approach in this case is to treat any cross between a public and private feature as private, and use it only in the user encoder.

The AM-SSP method has several advantages in practice. It is specifically designed to take advantage of public features, it is computationally cheaper in many cases, it preserves gradient sparsity, and the fact that we can reuse noised statistics (see Remark~\ref{rem:reusing_noise}) seems to further improve quality.

One limitation of the SSP approach is that the number of statistics to protect increases with the number of items. The method will generally work better when there are fewer items (and more examples per item). When too many fine-grained public features are used, this may lead to an increase in the number of unique items, and a degradation of quality. To give a hypothetical example, suppose font size is used as an ad feature. Then ads that are otherwise identical except for the font size will be treated as different items. We recommend excluding such fine-grained features from the public encoder to control the number of unique items (these excluded features can be incorporated in the private encoder). An interesting open question is to develop a more systematic approach to learning items that are \emph{similar but not identical} (in the sense that their public feature vectors are close in some metric). One promising observation is that since the feature matrix $\Xpub$ is public, one can compute such feature-based similarity without paying a privacy cost.

Another direction is motivated by our empirical observation that the correlated noise version of SSP performs better in practice than the independent noise version. Developing an analytical understanding of this variant, and conditions under which it provably improves utility, is an interesting open question.

% ======================================================================
\bibliography{refs}

% ==================================================================
\newpage
\appendix
\onecolumn
\section{Notation}
\label{app:notation}

We summarize notation used throughout the paper in this section for easy reference.

For a positive integer $n$, we denote by $[n]$ the set $\{1, \dots, n\}$. Given a convex set $\Ccal$, we denote by $\Pi_\Ccal$ the Euclidean projection on $\Ccal$ and by $\|\Ccal\|$ the diameter of $\Ccal$. For a vector $x$, we denote by $\clip(x, \Gamma)$ the Euclidean projection of $x$ on the L2 ball of radius~$\Gamma$.

The data is given by
\[
\Dcal = \{(\xpriv_i, y_i, j_i, k_i) \in \Rbb^{q} \times \Rbb \times  [m]\times[n]\}_{i \in \{1, \dots, \datasize\}},
\]
and we will always use $i$ to index training examples, $j$ to index items (i.e. rows in the public matrix $\Xpub$), and $k$ to index users (for user-level DP). We also use the following partitions of the data set: $\Dcal = \sqcup_{j = 1}^m \Oj = \sqcup_{k = 1}^n \Ok$, where $\Oj = \{i : j_i = j\}$ is the set of examples that involve public item $j$, and $\Ok = \{i : k_i = k\}$ is the set of examples that belong to user $k$.

The model is given by
\[
f_{\theta_u, \theta_v}(\xpriv_i, \xpub_j) = v_{\theta_v}(\xpub_j) \cdot v_{\theta_u}(\xpriv_i),
\]
and we use the shorthand $u_i(\theta_u) = u_{\theta_u}(\xpriv_i)$, and $v_j(\theta_v) = v_{\theta_v}(\xpub_{j_i})$. This allows to emphasize the dependence on the encoder's parameters.

Finally, the dimensions of the problem are summarized in the table below.

\begin{table}[h]
\centering
\begin{tabular}{c|p{7cm}}
\hline
$m$ & number of items \\
$n$ & number of users \\
$\datasize$ & number of examples \\
$p$ & dimension of public features ($\Xpub \in \Rbb^{m \times p}$) \\
$d$ & encoders' output dimension ($v_j(\theta_v) \in \Rbb^d$) \\
$d_v, d_u$ & encoders' parameters ($\theta_v \in \Rbb^{d_v}$)\\
\hline
\end{tabular}
\caption{Dimension Parameters}
\label{tab:my_label}
\end{table}

% --------------------------------------------------------------
\section{Proofs}
\label{app:proofs}
\subsection{Proof of Theorem~\ref{thm:priv_example}}
\begin{proof}
The result is an application of the Gaussian mechanism. Let $\Dcal, \Dcal'$ be two neighboring data sets that differ by a single example $(\xpriv_i, \xpub_{j_i}, y_i, j_i)$. This example only affects the statistics of item $j_i$ (all other statistics remain unchanged). Denote by $A_{j_i}, b_{j_i}$ (resp. $A'_{j_i}, b'_{j_i}$ the statistics computed on data sets $\Dcal$ (resp. $\Dcal'$). Then
\[
\|A_{j_i} - A'_{j_i}\|^2_F = \|\bar u_i \bar u_i^\top\|_F^2 \leq \Gamma_u^4,
\]
thus releasing $A_{j_i} + \sigma \Gamma_u^2 \Ncal^{d\times d}$ is $(\alpha, \frac{\alpha}{2 \sigma^2})$-RDP for all $\alpha > 1$. Similarly, we have
\[
\|b_{j_i} - b'_{j_i}\|^2_2 = \|\bar y_i \bar u_i\|_2^2 \leq \Gamma_u^2\Gamma_y^2,
\]
so releasing $b_{j_i} + \sigma \Gamma_u\Gamma_y \Ncal^{d\times d}$ is $(\alpha, \frac{\alpha}{2 \sigma^2})$-RDP.

By RDP composition, we have that Algorithm~\ref{alg:ssp} is $(\alpha, \alpha \frac{T\bar w^2}{\sigma^2})$-RDP (we release $2T$ noisy statistics), and Algorithm~\ref{alg:ssp-corr} is $(\alpha, \alpha \frac{\bar w^2}{\sigma^2})$-RDP (we release only 2 such statistics).

Translation from RDP to DP is standard, see for example~\citep[Proposition~3]{mironov2017renyi}: if a process is $(\alpha, \alpha\beta)$-RDP for all $\alpha$, and $\beta \leq \frac{\epsilon^2}{8\log 1/\delta}$, then the process is $(\epsilon, \delta)$-DP. The result follows from setting $\beta = \frac{T}{\sigma^2}$ for Algorithm~\ref{alg:ssp}, and $\beta = \frac{1}{\sigma^2}$ for  Algorithm~\ref{alg:ssp-corr}.
\end{proof}

% -----------------------------------------------------
\subsection{Proof of Theorem~\ref{thm:priv_user}}
\begin{proof}
Let $\Dcal, \Dcal'$ be two neighboring data sets that differ by a single user $k$. The user contributes to the sufficient statistics of all items $\{j_i, i \in \Ok\}$ (by definition of $\Ok$). Define $A$ to be the matrix $[A_1| \dots | A_m] \in \Rbb^{d \times md}$, and let $A'$ be the same matrix computed on $\Dcal'$ instead of $\Dcal$. Then
\[
\|A - A'\|_F^2 = \sum_{i \in \Ok} \|w_i \bar u_i \bar u_i^\top\|_F^2 \leq  \sum_{i \in \Ok} w_i^2 \Gamma_u^4 \leq \bar w^2 \Gamma_u^4
\]
where we used the definition of $\bar w^2 = \max_k \sum_{i \in \Ok} w_i^2$. Similarly, defining $b = [b_1, \dots, b_m] \in \Rbb^{d \times m}$ we have that
\[
\|b - b'\|_F^2 = \sum_{i \in \Ok} \|w_i \bar u_iy_i\|_2^2 \leq  \sum_{i \in \Ok} w_i^2 \Gamma_u^2\Gamma_y^2 \leq \bar w^2 \Gamma_u^2\Gamma_y^2.
\]
Therefore, releasing $A + \Gamma_u^2\sigma^2\Ncal^{d \times md}$ is $(\alpha, \frac{\alpha\bar w^2}{2 \sigma^2})$-RDP, and so is releasing $b + \sigma \Gamma_u\Gamma_y \Ncal^{d \times m}$.

The rest of the proof proceeds as in the proof of Theorem~\ref{thm:priv_example}.
\end{proof}

% ---------------------------------------------------------
\subsection{Proof of Theorem~\ref{thm:utility}}
\begin{proof}
We will use the following standard lemma, which can be found for example in~\citep[Lemma~2.5]{bassily2014erm}.
\begin{lemma}
\label{lemma:sgd}
Let $\Lcal$ be a convex function defined on a bounded domain $\Theta$, and let $\theta_v^* = \argmin_{\theta_v \in \Theta} \Lcal(\theta_v)$. Consider the SGD algorithm
\[
\theta_v^{t+1} = \proj_{\Theta}\left[\theta_v^t - \eta_t\hat g^t\right],
\]
where $\hat g^t$ satisfy the following: $\Exp[\hat g^t] = \nabla \Lcal(\theta_v^t)$, and $\Exp\|\hat g^t\|_2^2 \leq \Gcal^2$. Let $\eta_t = \frac{|\Theta|}{\Gcal\sqrt t}$. Then for all $t$,
\[
\Exp[\Lcal(\theta_v^t)] - \Lcal(\theta_v^*) = \Ocal\left(\frac{|\Theta|\Gcal \log t}{\sqrt t}\right).
\]
\end{lemma}
To apply the lemma, we will show that $\hat G^t$ (Line~\ref{line:ssp1_grad} in Algorithm~\ref{alg:ssp}) is an unbiased estimate of the gradient and bound its second moment. To make the notation concise, we will write $v_j^t = v_{\theta_v^t}(\xpub_j)$. Recall that the true gradient is
\begin{align*}
\nabla \Lcal(\theta_v)
= \sum_{j = 1}^m \xpub_j(A_j v_j - b_j)^\top
= \sum_{j = 1}^m \xpub_j\rho_j^\top
\end{align*}
where we defined $\rho_j = A_j v_j - b_j$. And by definition of the SSP1 algorithm, the gradient estimate is
\begin{align*}
\hat G^t 
&= \sum_{j = 1}^m \xpub_j [\hat A_j^t v_j^t - \hat b_j^t]^\top\\
&= \sum_{j = 1}^m \xpub_j [(A_j + \sigma \Gamma_u^2 \Ncal^{d \times d}) v_j^t - (b_j + \sigma \Gamma_y\Gamma_u\Ncal^d]^\top \\
&= \nabla \Lcal(\theta_v^t) + \sigma \sum_{j = 1}^m \xpub_j [\Gamma_u^2 \Ncal^{d \times d} v_j^t - \Gamma_y\Gamma_u\Ncal^d]^\top,
\end{align*}
where, importantly, the noise samples $\Ncal^{d\times d}, \Ncal^{d}$ are conditionally independent of the trajectory up to step $t$. Thus, it follows by independence that $\Exp[\hat G^t] = \nabla \Lcal(\theta_v^t)$, and we can bound the second moment
\begin{align}
\Exp[\|\hat G^t\|_2^2]
= \Exp[\|\nabla \Lcal(\theta_v^t)\|_2^2]
+ \sigma^2 \sum_{j = 1}^m \Exp[\|\xpub_j [\Gamma_u^2 \Ncal^{d \times d} v_j^t - \Gamma_y\Gamma_u\Ncal^d]^\top\|_2^2]\label{eq:proof_ghat}
\end{align}
(the cross-terms vanish by independence). It remains to bound individual terms. First, observe that
\[
\|\rho_j\|_2 = \|A_jv_j - b_j\|_2 \leq \|A_j\|_2 \|v_j\|_2 + \|b_j\|_2,
\]
where $\|A_j\|_2 = \|\sum_{i \in \Oj} u_iu_i^\top\|_2 \leq |\Oj|\Gamma_u^2$, $\|v_j\|_2 \leq \Gamma_y/\Gamma_u$ (by Assumption~\ref{assump:1}), and $\|b_j\|_2 = \|\sum_{i \in \Oj} y_iu_i\|_2 \leq |\Oj| \Gamma_y\Gamma_u$. Thus $\|\rho_j\|_2 \leq 2|\Oj|\Gamma_y\Gamma_u$, and
\begin{align}
\|\nabla \Lcal(\theta_v^t)\|_2
&\leq \sum_{j = 1}^m \|\xpub_j \rho_j^\top\|_2 \leq \sum_{j = 1}^m \|\xpub_j\|_2 \|\rho_j\|_2 \notag\\
&\leq\sum_{j = 1}^m 2\Gamma_x |\Omega_j|\Gamma_y\Gamma_u \notag\\
&= 2\datasize\Gamma_x\Gamma_y\Gamma_u \label{eq:proof_b1}
\end{align}
where we used that $\sum_j |\Oj| = \datasize$. This gives a bound on the first term in~\eqref{eq:proof_ghat}.

Turning to the second term in~\eqref{eq:proof_ghat}, we have that $\Exp[\|\Ncal^{d\times d}\|_2] = \sqrt d$ (induced norm) and $\Exp[\|\Ncal^d\|_2] = \sqrt d$ (Euclidean norm), thus
\begin{equation}
\Exp[\|\xpub_j[\Gamma_u^2 \Ncal^{d \times d} v_j^t - \Gamma_y\Gamma_u\Ncal^d]^\top\|_F] \leq 2 \Gamma_x\Gamma_y\Gamma_u\sqrt d.\label{eq:proof_b2}
\end{equation}
Combining bounds~\eqref{eq:proof_b1}-\eqref{eq:proof_b2} into~\eqref{eq:proof_ghat}, we get
\[
\Exp[\|\hat G^t\|_2^2] \leq [2\Gamma_x\Gamma_y\Gamma_u]^2(\datasize^2 + \sigma^2 md)
\]
Next, taking $\sigma^2 = \rho^2T$ and applying Lemma~\ref{lemma:sgd} with $\Gcal = 2\Gamma_x\Gamma_y\Gamma_u\sqrt{D^2 + mdT\rho^2}$, we get
\[
\Exp[\Lcal(\theta_v^T)] - \Lcal(\theta_v^*) = \tilde\Ocal\left(|\Theta|2\Gamma_x\Gamma_y\Gamma_u\sqrt{\frac{D^2}{T} + md\rho^2}\right).
\]
Finally we choose $T$ to equalize the two terms, $T = \frac{D^2}{md\rho^2}$ (under this choice of $T$, $\Gcal = 2\sqrt2\Gamma_x\Gamma_y\Gamma_u \datasize$), to obtain the final desired bound $\Ocal\left(2|\Theta|\Gamma_x\Gamma_y\Gamma_u\rho\sqrt{2md}\right)$.
\end{proof}

% -------------------------------------------------------------
\section{Batched Algorithms and Complexity Analysis}
\label{app:complexity}
In this section, we give additional details related to the batched variants of our algorithms, and discuss computational complexity. 

The mini-batched versions of SSP1 and SSP2 are given respectively in Algorithms~\ref{alg:mini-ssp} and~\ref{alg:mini-ssp-corr}, where we highlight differences compared to the full-batch case in blue.

For mini-batch DP-SGD, notice that if several examples in a batch have the same feature vector $\xpub_j$, then the forward/backward passes $v_j = v_{\theta_v}(\xpub_j)$ and $\Jcal_j = \frac{\partial v_{\theta_v}(\xpub_j)}{\partial \theta_v}$ only have to be computed once. This is summarized in Algorithm~\ref{alg:dpsgd}.

\begin{algorithm}[h!]
\caption{Mini-batch SSP1}
\label{alg:mini-ssp}
\SetAlgoLined
\DontPrintSemicolon
\SetKwProg{proc}{Procedure}{}{}
{\bf Inputs}: Public features $\Xpub$, training data $\Dcal = \{(\xpriv_i, y_i, j_i, k_i)\}_{i \in \{1, \dots, \datasize\}}$, optional weights $\{w_i\}$, number of steps $T^D$, clipping parameters $\Gamma_y, \Gamma_u$, noise standard deviation $\sigma$, learning rate $\eta_t$, initial parameters $\theta_v^{0}, \theta_u^0$.\\
Let $\bar u_i = \clip(u_{\theta_u}(\xpriv_i), \Gamma_u)$,
$\bar y_{i} = \clip(y_{i}, \Gamma_y)$\\
\For{$0 \leq t \leq T^D-1$}{
\blue{
Uniformly sample $B \subset [\datasize]$\\
\For{$1 \leq j \leq m$}{
$\hat A^t_j = \sum_{i \in \Oj \cap B} w_i\bar u_i \bar u_i^\top + \sigma\Gamma_u^2 \Ncal^{d \times d}$\\
$\hat b^t_j = \sum_{i \in \Oj \cap B} w_i\bar y_i \bar u_i + \sigma\Gamma_y\Gamma_u \Ncal^{d}$\\
}
}
$\hat G^t \leftarrow \sum_{j = 1}^m \frac{\partial v_{\theta_v^t}(\xpub_j)}{\partial \theta_v}(\hat A_j^t v_{\theta_v^t}(\xpub_j) - \hat b_j^t)$\\
$\theta_v^{t+1} \leftarrow \theta_v^t - \eta_t \hat G^t$
}
\KwRet $\theta_v^T$
\end{algorithm}

% -------------------------------------------------------------

\begin{algorithm}[h!]
\caption{Mini-batch SSP2}
\label{alg:mini-ssp-corr}
\SetAlgoLined
\DontPrintSemicolon
\SetKwProg{proc}{Procedure}{}{}
{\bf Inputs}: Public features $\Xpub$, training data $\Dcal = \{(\xpriv_i, y_i, j_i, k_i)\}_{i \in \{1, \dots, \datasize\}}$, optional weights $\{w_i\}$, number of steps $T^I$, clipping parameters $\Gamma_y, \Gamma_u$, noise standard deviation $\sigma$, learning rate $\eta_t$, initial parameters $\theta_v^{0}, \theta_u^0$.\\
Let $\bar u_i = \clip(u_{\theta_u}(\xpriv_i), \Gamma_u)$,
$\bar y_{i} = \clip(y_{i}, \Gamma_y)$\\
\For{$1 \leq j \leq m$}{
$\hat A_j \leftarrow \sum_{i \in \Oj} w_i\bar u_i \bar u_i^\top + \sigma\Gamma_u^2 \Ncal^{d \times d}$\\
$\hat b_j \leftarrow \sum_{i \in \Oj} w_i\bar y_i \bar u_i + \sigma\Gamma_y\Gamma_u \Ncal^d$.
}
\For{$0 \leq t \leq T^I-1$}{
\blue{Uniformly sample $B \subset [m]$}\\
$\hat G^t \leftarrow \sum_{\blue{j \in B}} \frac{\partial v_{\theta_v^t}(\xpub_j)}{\partial \theta_v}(\hat A_j v_{\theta_v^t}(\xpub_j) - \hat b_j)$\\
$\theta_v^{t+1} \leftarrow \theta_v^t - \eta_t \hat G^t$
}
\KwRet $\theta_v^T$
\end{algorithm}

% -------------------------------------------------------------
\begin{algorithm}[h!]
\caption{Mini-batch DP-SGD (adapted to multi-encoder models for improved efficiency)}
\label{alg:dpsgd}
\SetAlgoLined
\DontPrintSemicolon
\SetKwProg{proc}{Procedure}{}{}
{\bf Inputs}: Public features $\Xpub$, training data $\Dcal = \{(\xpriv_i, y_i, j_i, k_i)\}_{i \in \{1, \dots, \datasize\}}$, optional weights $\{w_i\}$, number of steps $T^D$, clipping parameters $\Gamma_g$, noise standard deviation $\sigma$, learning rate $\eta_t$, initial parameters $\theta_v^{0}, \theta_u^0$.\\
Let $u_i = u_{\theta_u^0}(\xpriv_i)$\\
\For{$0 \leq t \leq T^D-1$}{
\blue{
Uniformly sample a batch $B \subset [\datasize]$.\\
\For{$j \in \{j_i : i \in B\}$}{
$v_j \leftarrow v_{\theta_v}(\xpub_j)$\\
$\Jcal_j \leftarrow \frac{\partial v_{\theta_v}(\xpub_j)}{\partial \theta_v}$
}
}
$\hat G^t \leftarrow \displaystyle \sum_{\blue{i \in B}} \clip\left(\Jcal_{j_i}(u_i^T v_{j_i} - y_i)u_i, \Gamma_g\right) + \sigma \Gamma \Ncal^{d_v}$\\
$\theta_v^{t+1} \leftarrow \theta_v^t - \eta_t \hat G^t$
}
\KwRet $\theta_v^T$
\end{algorithm}
% -------------------------------------------------------------

\newpage
\paragraph{Proof of Proposition~\ref{prop:complexity}} We now turn to the complexity analysis of mini-batch SSP2 and DP-SGD.

\begin{proof}
Let $c$ be the cost of computing one forward and one backward pass, i.e. the cost of computing $v_j$ and $\Jcal_j$ for one item $j$. We also recall the following quantities: $\beta_j$ is the expected number of batches that contains item $j$ in one epoch, and $\beta = \sum_{j = 1}^m \beta_j$.

First, consider DP-SGD. For some item $j$, every time the item is visited, we compute one forward/backward pass for that item (Lines 6-7), for a cost of $c\sum_{j = 1}^m \beta_j = \beta$, hence the cost of DP-SGD over $e$ epochs is at least $\Omega(ce\beta)$. Note that there is the additional cost of computing and clipping the gradients (Line 8), but for the purposes of our analysis, we only need a lower bound on the cost of DP-SGD (we will compare an upper bound on the cost of SSP to this lower bound on the cost of DP-SGD).

We now consider SSP2. First, there is the cost of computing the sufficient statistics (Lines 3-5). This can be done by iterating over all examples in $\Dcal$ and accumulating the statistics. This costs $\Ocal(\datasize (d^2 + d))$ ($d^2$ for accumulating the $\hat A_j$'s, and $d$ for accumulating the $\hat b_j$'s). Second, for each batch, we compute the gradient $\hat G^t$ (Line 8). This consists of the following operations for each $j \in B$: computing the forward/backward passes $v_j, \Jcal_j$ (a cost of $c$), then computing the vector $\hat A_j v_j - \hat b_j$ (a cost of $d^2$), and finally multiplying this vector by $\Jcal_j$ (a cost of at most $c$, because the cost of computing the Jacobian is greater than the cost of multiplying by the Jacobian). The total cost for item $j$ is therefore $\Ocal(c + d^2)$. Finally, the total cost over $e$ epochs $\Ocal(\datasize d^2 + em(c+d^2))$ (the first term is the cost of computing statistics, and the second is the cost of computing gradients from the statistics). This completes the proof.
\end{proof}

\paragraph{Comparison of Computational Complexity} We expand on the complexity discussion of Section~\ref{sec:complexity}. The ratio between the cost of SSP2 and the cost of DP-SGD (see Proposition~\ref{prop:complexity}) is bounded by
\begin{equation}
\label{eq:app:ratio}
\Ocal\left(\frac{\datasize d^2 + em(c+d^2)}{ce\beta}\right) = \Ocal\left(\frac{m}{\beta}\Big(1+\frac{d^2}{c} + \frac{\datasize d^2}{cem}\Big)\right)    
\end{equation}

This depends on several problem parameters:
\begin{itemize}[leftmargin=13pt,topsep=0pt,itemsep=0ex,partopsep=0ex,parsep=1pt]
\item $\beta/m$: this represents the average number of visits per item: the larger this number is, the bigger the advantage of SSP. \item $\frac{d^2}{c}$: this term depends on the encoder's architecture. $c$ is the cost of one forward/backward pass, and $d$ is the output dimension of the encoder.
\item the last term $\frac{\datasize d^2}{cem}$ represents the relative overhead of computing statistics in SSP, and decreases with the number of epochs~$e$ (if the number of epochs is large enough, the overhead is amortized).
\end{itemize}

\begin{figure}[ht]
    \centering
    \includegraphics[width=.40\textwidth]{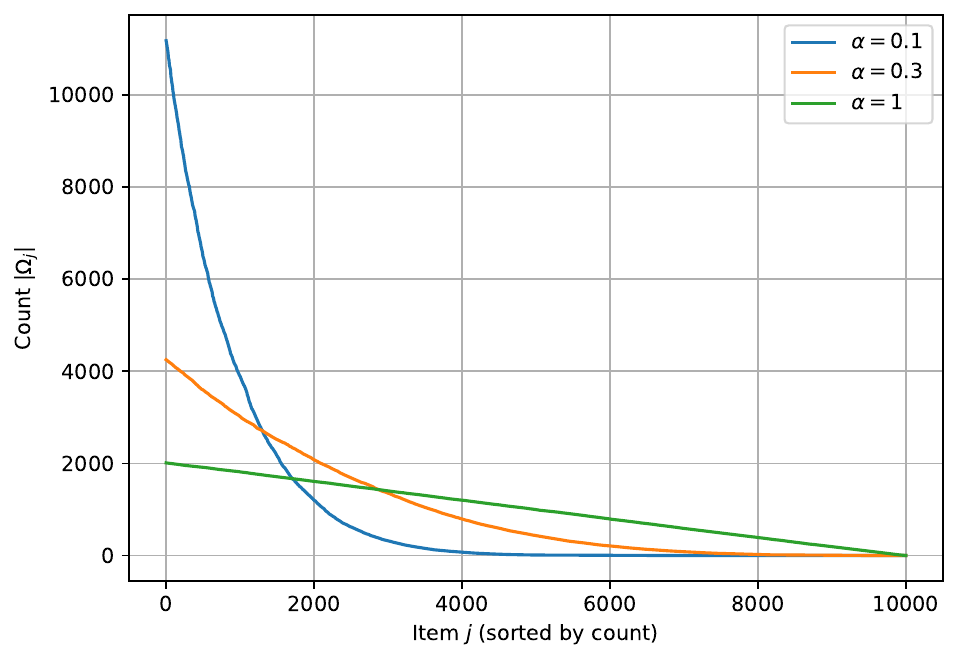}
    \hspace{.1\textwidth}
    \includegraphics[width=.40\textwidth]{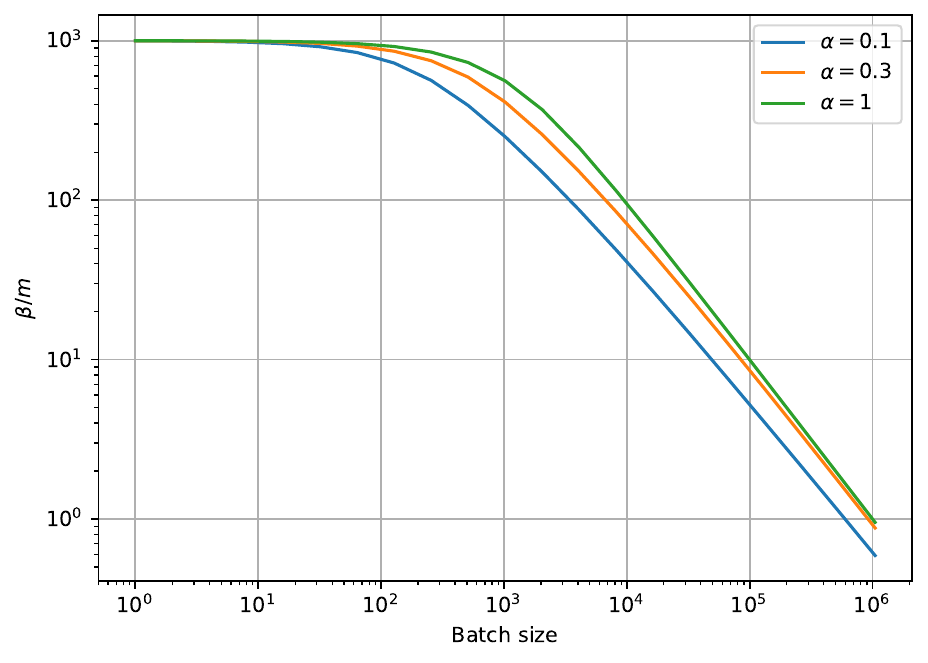}
    \caption{Item visits under a power law distribution. In this example the data set size is $\datasize = 10^6$ and the number of items is $m = 10^3$. The distribution of item counts follows a power law distribution with density $f(x) \propto x^{-\alpha}$ where $\alpha$ is a positive parameter. The left plot shows the count distribution, and the right plot shows the average number of visits $\beta/m$ as we vary the batch size $B$.}
    \label{fig:item_visits}
\end{figure}

First, consider the term $\beta/m$. Assuming that examples are sampled independently and with replacement, we can get a precise estimate of $\beta$: let $p_j = \frac{|\Oj|}{\datasize}$ be the probability of sampling item $j$. For a batch size $B$, the probability that item $j$ appears (at least once) in the batch is $q_j = 1 - (1-p_j)^B$. The number of batches in which item $j$ appears is a Binomial distribution with probability $q_j$, and since there are $\datasize/B$ batches in one epoch, the expected number of batches containing $j$ in one epoch is $\beta_j = \datasize/B(1 - (1-p_j)^B)$. This is entirely determined by the item frequencies $p_j$, the data size $\datasize$, and the batch size $B$. We plot a few examples in Figure~\ref{fig:item_visits}, where we take the item count distribution to follow a power-law distribution, commonly encountered in practice in recommender systems and ads applications~\citep{yin2012longtail,liu2020longtail}. The plot shows that as the batch size increases, the total number of visits $\beta$ decreases. When $B = 1$, $\beta = D$, and when $B = \datasize$ (full batch), $\beta/m \approx 1$ (if the batch is all of $\Dcal$, then $\beta = m$, but since we are sampling with replacement, some items may not be sampled at all so one could have $\beta$ slightly lower than $m$ as can be seen on the plot). In the example, we take $\datasize = 10^6$ and $m = 10^3$. Notice that $\beta/m$ remains large (more than a hundred) for batch sizes up to a few thousands.

\begin{figure}[h]
\centering
\begin{subfigure}[b]{0.40\textwidth}
\centering
\includegraphics[width=\textwidth]{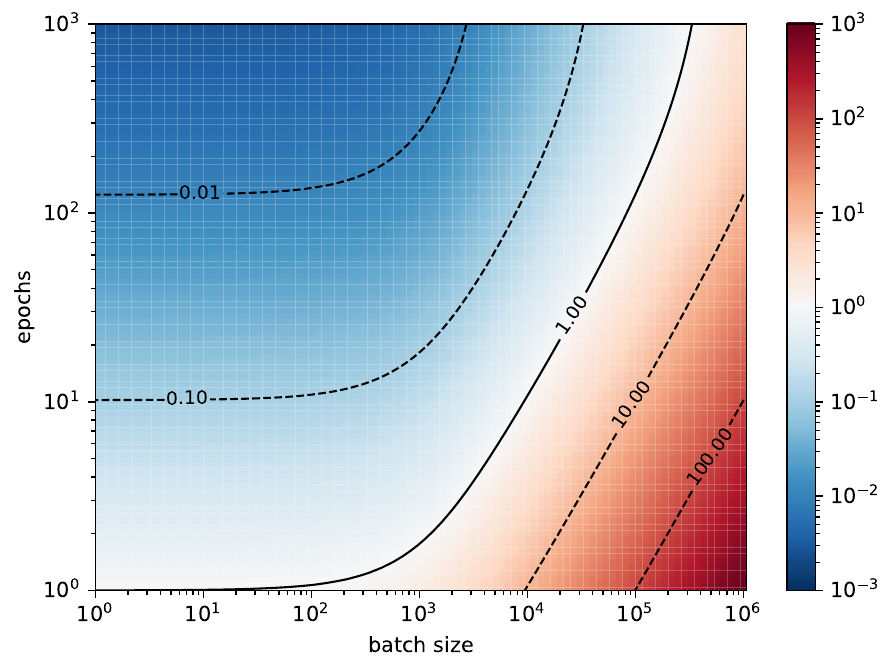} \caption{$c=d^2$}
\end{subfigure}%
\hspace{.1\textwidth}%
\begin{subfigure}[b]{0.40\textwidth}
\centering
\includegraphics[width=\textwidth]{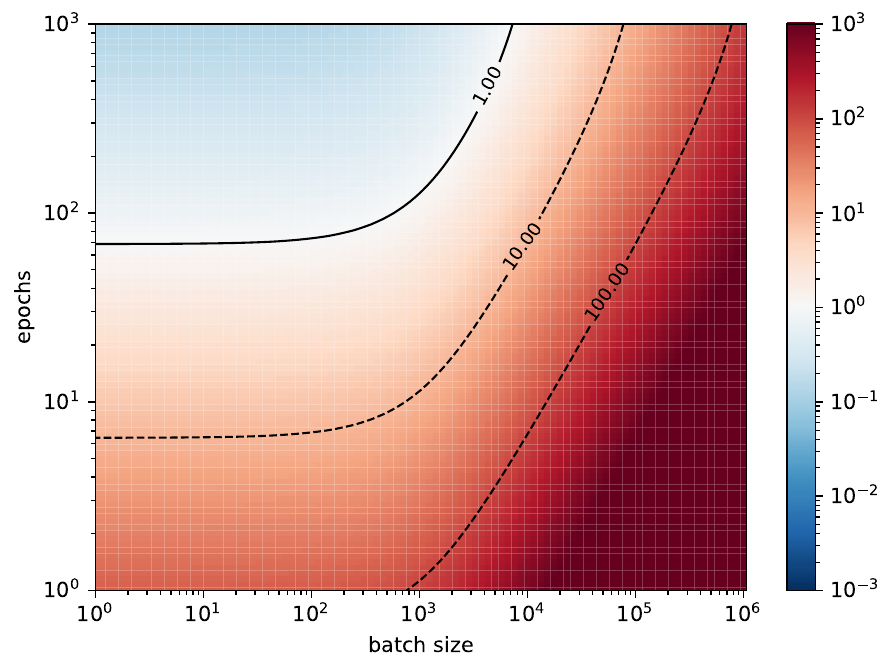}
\caption{$c=d$}
\end{subfigure}
\caption{Complexity ratio (cost of SSP divided by cost of DP-SGD) as we vary the batch size and number of epochs. In this example, $\datasize=10^6, m=10^3$. Two representative cases are shown: $c = d^2$ corresponding to encoders with at least one hidden layer (left), and $c = d$, corresponding to id-only linear encoders (right). The solid line shows the case where both algorithms have comparable cost.}
\label{fig:complexity}
\end{figure}

Once $\beta/m$ is estimated, we can easily compute the ratio~\eqref{eq:app:ratio}. In Figure~\ref{fig:complexity}, we plot the ratio~\eqref{eq:app:ratio} as we vary the batch size and the number of epochs. We fix the data size $\datasize = 10^6$ (changing this parameter will change the scale the graph, but the trends will be similar).

As discussed above, the ratio $\frac{d^2}{c}$ depends on the model architecture, we consider two representative cases: the first is when the encoder contains at least a hidden layer of width larger than $d$, in which case we have $c \geq d^2$ (whenever the encoder has a hidden layer, its width is usually more than the width of the last layer). The second case is when the encoder is a single linear layer, in which case $c = kd$ where $k$ is the number of non-zero features per item. In the simplest case (the most favorable to DP-SGD), there is a single active feature per item and $c = d$.

In the case $c = d^2$, there a large region in which SSP has a computational advantage, and the advantage increases as the number of epochs increases and as the batch size decreases. The advantage can easily be several orders of magnitude in the small batch regime (batch size of 100-1000).

The case $c = d$ is intuitively the least favorable to SSP, since the cost of computing the statistics ($d^2$) is much higher than the cost of the encoder ($c = d$), so it takes longer to amortize the cost of computing the statistics. In this case, we expect SSP to be more expensive, unless the number or epochs is very large.

% =============================================================
\section{SSP Algorithm with Non-Quadratic Losses}
\label{app:convex}
In this section, we discuss a generalization of our algorithms from the quadratic to the convex loss case. The main idea is to compute and optimize successive quadratic approximations of the loss, much like in second-order methods. The main difference is that instead of computing the approximation in the $\theta_v$ space (which would be intractable in most practical settings), we compute it in the $v$ space, i.e. a quadratic approximation w.r.t. \emph{the encoder's output}.

More precisely, consider the loss~\eqref{eq:loss}, reproduced below
\[
\Lcal(\theta_v) = \sum_{i = 1}^D \ell\left(u_i \cdot v_{j_i}(\theta_v), y_i\right),
\]
where $\ell: \Rbb \times \Rbb \to \Rbb$ is convex (note however that $\Lcal$ is not, since the item encoder $v_{\theta_v}$ is typically non-linear). Writing $\ell_i(v) = \ell(u_i \cdot v, y_i)$, the loss $\Lcal$ is the composition $\Lcal(\theta_v) = \sum_{i = 1}^\datasize \ell_i(v_{j_i}(\theta_v))$.

By the chain rule, the gradient and Hessian of $\ell_i$ at $v^0$ are given respectively by $\nabla \ell_i(v^0) = \ell'(u_i\cdot v^0, y_i) u_i$ and $\nabla^2\ell_i(v^0) = \ell''(u_i\cdot v^0, y_i) u_i u_i^\top$, where $\ell'$ and $\ell''$ denote the first and second derivatives of $\ell$ w.r.t. its first argument. To simplify notation, let us write $\ell_i(v^0) = \ell(u_i\cdot v^0, y_i)$, $d\ell_i(v^0) = \ell'(u_i\cdot v^0, y_i)$ and $d^2\ell_i(v^0) = \ell''(u_i\cdot v^0, y_i)$.

Now, a simple Taylor expansion of $\ell_i$ around $v^0$ yields the quadratic approximation
\[
\tilde \ell_i(v) = \ell_i(v^0) + d\ell_i(v^0) u_i\cdot (v - v^0) + \\
\frac{1}{2} d^2\ell_i(v^0) (v - v^0)^\top u_i u_i^\top (v - v^0).
\]
Denoting $v_j^0 = v_j(\theta_{v}^0)$, we obtain the following approximation of $\Lcal$ around $\theta_v^0$:
\begin{equation}
\label{eq:app:loss_approx}
\tilde \Lcal(\theta_v) = \sum_{j = 1}^m c_j +  g_j^\top(v_j(\theta_v) - v_j^0) + \\
\frac{1}{2} (v_j(\theta_v) - v_j^0)^\top H_j (v_j(\theta_v) - v_j^0),
\end{equation}
where
\begin{align}
c_j &= \sum_{i \in \Oj} \ell_i(v^0) \notag\\
g_j &= \sum_{i \in \Oj} d\ell_i(v_j^0) u_i \notag\\
H_j &= \sum_{i \in \Oj} d^2\ell_i(v_j^0) u_iu_i^\top \label{eq:app:stats}
\end{align}
Since this is a quadratic function in $v$, this gives us a similar gradient decomposition, which we state below.
\begin{proposition}
The gradient of the quadratic approximation~\eqref{eq:app:loss_approx} is given by
\begin{equation}
\label{eq:app:grad}
\nabla \tilde\Lcal(\theta_v) = \sum_{j = 1}^m \Jcal_j(\theta_v)[H_j (v_j(\theta_v) - v_j^0) + g_j],
\end{equation}
where we use the shorthands $\Jcal_j(\theta_v) = \frac{\partial v_{\theta_v}(\xpub_j)}{\partial \theta_v}$ and $v_j(\theta_v) = v_{\theta_v}(\xpub_j)$.
\end{proposition}
As in Proposition~\ref{prop:decomposition}, the terms $\Jcal_j$ and $v_j$ only depend on public data and don't need protection. The only difference is in the sufficient statistics: Instead of $A_j, b_j$, the decomposition now involves $H_j, g_j$. This motivates Algorithm~\ref{alg:convex-ssp}. The main difference with Algorithm~\ref{alg:ssp} is that $H_j, g_j$ depend on where we take the approximation (note the dependence on $v^0$ in~\eqref{eq:app:stats}), so they need to be periodically recomputed.

\begin{algorithm}[h]
\caption{SSP-convex: Sufficient Statistics Perturbation for Convex Losses (with correlated noise)}
\label{alg:convex-ssp}
\SetAlgoLined
\DontPrintSemicolon
\SetNoFillComment
\SetKwProg{proc}{Procedure}{}{}
{\bf Inputs}: Public features $\Xpub$, training data $\Dcal = \{(\xpriv_i, y_i, j_i, k_i)\}_{i \in \{1, \dots, \datasize\}}$, optional weights $\{w_i\}$, number of steps $T$, number of inner iterations $\tau^{\max}$, clipping parameters $\Gamma_H, \Gamma_g$, noise standard deviation $\sigma$, learning rate $\eta$, initial parameters $\theta_v^{0}, \theta_u^{0}$.\\
Let $\bar u_i = \clip(u_{\theta_u}(\xpriv_i), \Gamma_u)$,
$\bar y_{i} = \clip(y_{i}, \Gamma_y)$\\
\For{$0 \leq t \leq T-1$}{
\For{$1 \leq j \leq m$}{
\tcc*[l]{Compute statistics for the quadratic approximation around $v^{t}$}
$\hat H_j^t \leftarrow \sum_{i \in \Oj} w_i \clip(d^2\ell_i(v_j^t) \bar u_i \bar u_i^\top, \Gamma_H) + \sigma\Gamma_H \Ncal^{d \times d}$\\
$\hat g_j^t \leftarrow \sum_{i \in \Oj} w_i\clip(d\ell_i(v_j^t) \bar u_i, \Gamma_g) + \sigma\Gamma_g \Ncal^d$.
}
$\theta_v^{(0)} \leftarrow \theta_v^t$\\
\For{$0 \leq \tau \leq \tau^{\max}-1$}{
\tcc*[l]{Optimize the quadratic approximation}
$\hat G^{(\tau)} \leftarrow \sum_{j = 1}^m \Jcal_j(\theta_v^{(\tau)})[\hat H_j^t (v_j(\theta_v^{(\tau)}) - v_j(\theta_v^{t})) + \hat g_j^t]$\\
$\theta_v^{(\tau+1)} \leftarrow \theta_v^{(\tau)} - \eta \hat G^{(\tau)}$
}
$\theta_v^{t+1} \leftarrow \theta_v^{(\tau^{\max})}$\\
}
\KwRet $\theta_v^{T}$
\end{algorithm}

Algorithm~\ref{alg:convex-ssp} consists of multiple calls to SSP2 (Algorithm~\ref{alg:ssp-corr}), each applied to the quadratic approximation around the current iterate $v^t$.
Note that the main iterates are denoted by $\theta_v^t$, while the inner loop uses $\theta_v^{(\tau)}$.

The privacy guarantee is an immediate extension of Theorems~\ref{thm:priv_example}-\ref{thm:priv_user}: to guarantee $(\epsilon, \delta)$-DP, it suffices to take $\sigma = \frac{\sqrt{8T\log 1/\delta}}{\epsilon}$ for example-level DP, and $\sigma = \frac{\bar w\sqrt{8T\log 1/\delta}}{\epsilon}$ for user-level DP. Notice that both scale with $\sqrt{T}$ (since sufficient statistics are recomputed at each step).

Empirical evaluation of Algorithm~\ref{alg:convex-ssp} is outside the scope of this paper, and is left as future work.

% --------------------------------------------------------------
\newpage
\section{Additional Experiments}
\label{app:exp}

In this section, we provide additional details regarding the exact experimental setup, along with additional results.

\subsection{Experimental Setup}
\label{app:datasets}

\paragraph{Data Sets and Public Features} We use two benchmarks based on MovieLens data\footnote{The license information can be found on the GroupLens website at the following URLs: \url{https://files.grouplens.org/datasets/movielens/ml-10m-README.html} and \url{https://files.grouplens.org/datasets/movielens/ml-20m-README.html}}. The first is a regression task proposed by~\cite{lee13llorma}, and the second is a classification task proposed by~\cite{liang18vae}.

The MovieLens data include movie features consisting of the release year and 19 movie genres. \cite{curmei2023private} expanded the movie features by extracting additional metadata from the Wikidata website; the additional features consist of a finer set of 310 genres, as well as 58,139 persons covering 55 roles. While the release year is a univalent feature, all other features (genres, persons, and roles) are multivalent features. Furthermore, roles and persons are paired, and one person may have multiple roles in a movie, and a role, such as ``actor", may be associated to multiple persons. In our experiments, we simply treat each feature category as a bag of words (where we count repetitions, so if a feature such as ``actor" is repeated, this will be treated as a weight of that feature value). One alternative is to use a feature cross between person and role, so each unique pair defines a feature value.

Statistics about the two data sets are summarized in Table~\ref{tab:stats}.

\begin{table}[h]
\centering
\begin{tabular}{cccc}
\toprule
data set & $n$ (users) & $m$ (items) & $p$ (features)\\
\midrule
ML10M &  69,878 & 10,677 & 58,619\\
ML20M & 136,677 & 2010 & 25,805\\
\bottomrule
\end{tabular}
\caption{Statistics of the MovieLens data sets}
\label{tab:stats}
\end{table}

\begin{figure}[h!]
\centering
\includegraphics[width=0.4\textwidth]{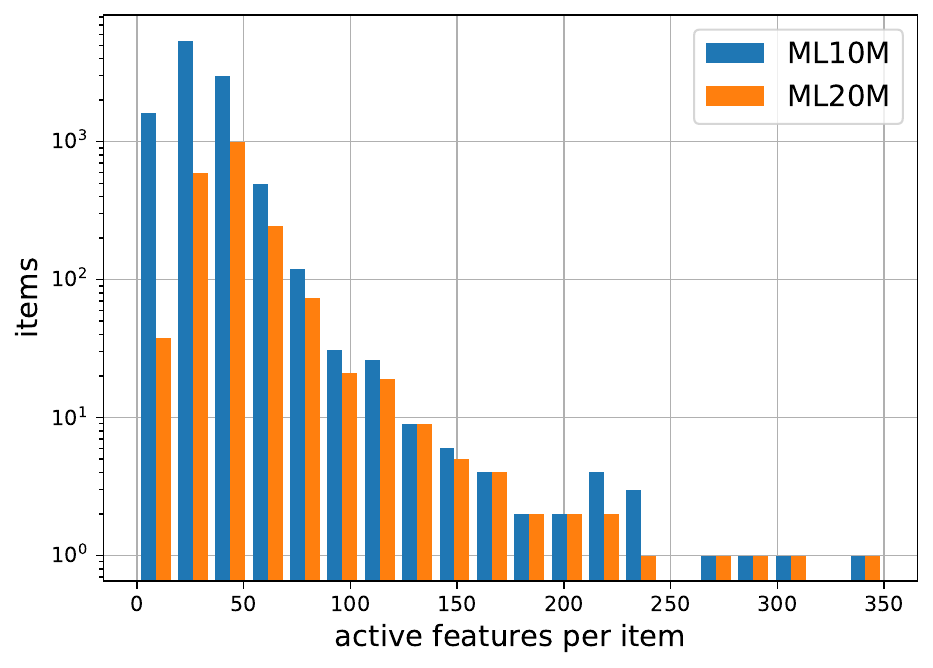}
\caption{Density of features across movies}
\label{fig:feature_density}
\end{figure}

Following previous work~\citep{jain2018differentially,chien2021private,krichene2023multi}, ML10M uses the entire data, while in ML20M, training is restricted to the top 10\% most frequent movies (but evaluation is always done on the entire set of movies).
The histograms of the number of features active for each item are reported in Figure~\ref{fig:feature_density}. The difference between the two distributions is explained by the fact that ML20M is restricted to the top frequent movies, for which metadata tends to be more comprehensive.

\paragraph{Model Architecture and Training} The public encoder consists of a two-layer model. The first layer is an embedding layer where each one of the 5 features (year, original genre, extended genre, person, role) is mapped to an embedding vector\footnote{For simplicity, we use the same embedding dimension $d$ for all features, and we also use the same dimension $d$ as the output dimension of the second layer.} in $\Rbb^d$ (multivalent features are treated as a bag of words and their embeddings are averaged). These 5 embeddings are concatenated to form the hidden layer in $\Rbb^{5d}$, and followed by a dense layer\footnote{Note that for this architecture, the cost $c$ in our complexity analysis would be approximately $5d^2$, since the dense layer is a matrix in $\Rbb^{5d \times d}$.} with output dimension $\Rbb^d$. In our experiments, the dimension $d$ is treated as a hyper-parameter, and we consider dimensions up to $d = 64$, following the DP-ALS and DP-CMF baselines we compare to. We observe that higher quality can be achieved on ML20M with higher dimension (but to make a fair comparison to the baselines, we restrict to $d \leq 64$).

As for the user encoder, we follow the same setup as~\citep{jain2018differentially,chien2021private,krichene2023multi,curmei2023private}. The user encoder is an embedding lookup, i.e. each user $k$ is mapped to a unique vector $u_k$ (there are no other user features except the id). The model's output is therefore $u_{k_i}\cdot v_{\theta_v}(\xpub_{j_i})$ ($k_i$ is the user id, $j_i$ is the movie id).

Each model is trained on an NVIDIA Tesla P100 GPU. The model is trained by minimizing the regularized quadratic loss
\[
\Lcal(\theta_v) = \frac{1}{2}\sum_{i = 1}^{\datasize} (u_i \cdot v_{j_i}(\theta_v) - y_i)^2 + \lambda_u \|u_i\|_2^2 + \lambda_v \|v_{j_i}(\theta_v)\|_2^2.
\]
We use the same \userUpdate{} for all methods (see footnote 2). During the \userUpdate{}, since we learn an embedding per user (with no shared parameters across users), the problem reduces to $n$ decoupled least squares problems (one per user):
\[
u_k^* = \argmin_{u} \frac{1}{2}\sum_{i \in \Ok} (u \cdot v_{j_i}(\theta_v) - y_i)^2 + \lambda_u \|u_k\|_2^2,
\]
for which we use the closed form solution $u_k^* = (\sum_{i \in \Ok}v_{j_i}v_{j_i}^\top + \lambda_u I)^{-1}(\sum_{i \in \Ok} y_i v_{j_i})$.

\paragraph{Evaluation}
\def\testset{\Dcal_{\text{test}}}
\def\Oktrain{\Ok_{\text{history}}}
\def\Oktest{\Ok_{\text{target}}}

We follow the protocol of the original benchmarks~\cite{lee13llorma,liang18vae}, which we summarize below. Each data set is split into training/validation/test.

In ML10M, the set of ratings is split at random, and utility is measured using RMSE, defined as
$\text{RMSE} = \sqrt{\frac{\sum_{i \in \testset} (u_i \cdot v_{j_i} - y_i)^2}{|\testset|}}$, where $\testset$ is the set of test ratings.

In ML20M, the validation and test splits consist of held-out users. Since validation/test users are not present in the training data, at evaluation time, one first needs to compute an embedding $u_k$ for those users to be able to generate predictions. For this purpose the benchmark also splits the examples of each validation/test user $k$ into an 80-20 split $\Oktrain \sqcup \Oktest$, where the first part is used to compute the user embedding (by running \userUpdate{} on $\Oktrain$), and the second is used as the target labels. Utility is measured using Recall@20, defined as follows. A prediction score is computed for all movies except in $\Oktrain$ (to avoid penalizing a model that recommends items from the user's history). Let $\hat\Omega^k$ be the set of top-20 predictions, then the recall is defined as
$\text{Recall@20} = \frac{1}{n}\sum_{k = 1}^n \frac{|\Oktest \cap \hat\Omega^k|}{\min(20, |\Oktest|)}$.

Hyper-parameters are tuned on the validation data and metrics are reported on test data. Plots in Figures~\ref{fig:ml10m},\ref{fig:ml20m},\ref{fig:comp} report the mean metric value across ten runs and the standard deviation of the metric value is depicted as a shaded region (small standard deviations are barely visible in some instances).

%While Recall@20 is the average recall across all users in the test set, where $\mbox{Recall@20}(k)$ of a user $k$ is the ratio computed as follows. First, note that the items in the user's history and the ones in the user's test are two disjoint sets. A prediction score is computed for all items other than the items in the user's history (to avoid penalizing a model that recommends items from the user's history), the numerator of $\mbox{Recall@20}(k)$ is the number of items in the test set with a score among the top 20 scored items; the denominator is the minimum between 20 and the number of items in the set for user $k$ (so that a perfect recall of 1 is always attainable).

\paragraph{Privacy Budget Allocation} All algorithms (including the DP-ALS and DP-CMF baselines) use the budget allocation method of~\cite{krichene2023multi}. The method consists of computing example weights $w_i$ (these are the inputs weights $w_i$ in Algorithms~\ref{alg:ssp}-\ref{alg:ssp-corr}) that are designed to control the sensitivity of each example. By allocating a higher weight (hence sensitivity) to tail items, it was found that overall utility can improve. The weights are computed following the method of~\citep[equations (11)-(12)]{krichene2023multi}, which we summarize here. First, compute a DP estimate of the item counts, $\hat n_j$, then define the weights $w_i = \bar w \frac{\hat n_{j_i}^{-1/4}}{\sqrt{\sum_{i \in \Ok} n_{j_i}^{-1/4}}}$, where $\bar w$ is a parameter that control the privacy guarantee. By definition, we have that $\sum_{i \in \Ok} w_i^2 = \bar w^2$, so this parameter corresponds to the $\bar w^2$ in Theorems~\ref{thm:priv_user}. Finally, privacy accounting is done via RDP composition~\citep{mironov2017renyi}, where we also account for the estimation of item counts described above. The accounting is identical across all algorithms we evaluated.

\begin{figure}[h!]
\centering
\includegraphics[width=0.40\textwidth]{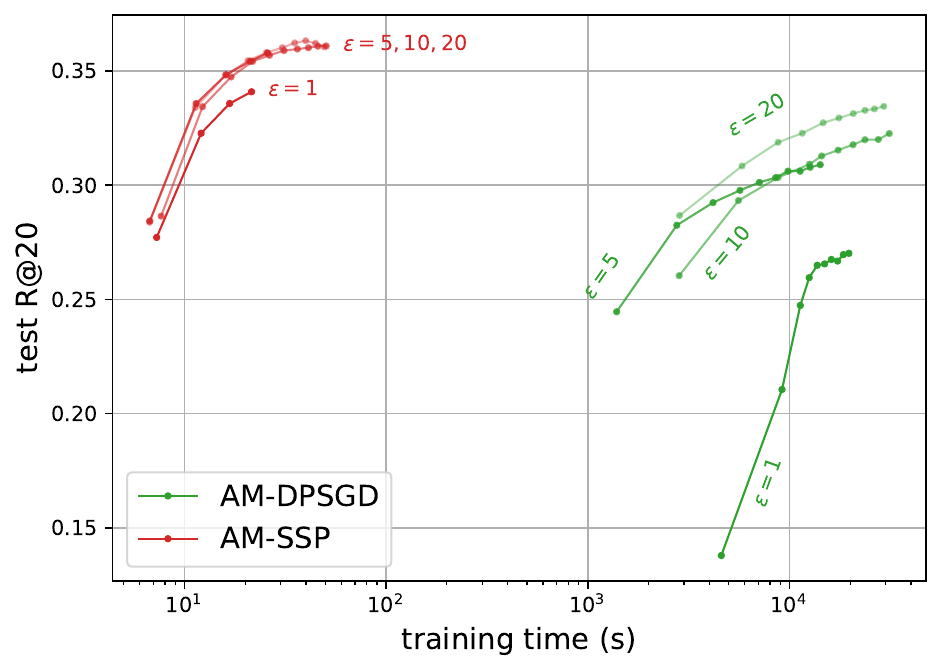}
\caption{Comparison of SSP1 and SSP2 on ML20M}
\label{fig:comp20m}
\end{figure}

\paragraph{The DP-SGD baseline} In addition to SOTA baselines (DP-ALS and DP-CMF), we also compare to plain DP-SGD, where all model parameters are optimized jointly (without alternating minimization). This can be considered a weaker baseline, since it is not adapted to the problem's structure. The DP-SGD result is reported in Figure~\ref{fig:ml10m}. It achieves a lower utility than all other methods, for example the RMSE at $\epsilon = 20$ is worse than the RMSE of competing methods at $\epsilon = 5$. Similarly for ML20M, the utility is much worse than other methods: we measured a Recall@20 of 0.206 at $\epsilon = 20$, while all other methods have better Recall@20 even at $\epsilon = 1$.

\subsection{Computational Cost on ML20M}

Figure~\ref{fig:comp20m} compares training a recall model with  AM-DPSGD and AM-SSP on the ML20M dataset. The overall observation is similar to the one drawn from Figure~\ref{fig:comp}, and in this case there is approximately three-orders of magnitude difference in training time between the two methods.

It's worth mentioning that we use an efficient implementation for DP-SGD (using the TensorFlow Privacy library\footnote{\url{www.tensorflow.org/responsible_ai/privacy}} that implements fast per-example gradient clipping~\citep{goodfellow2015efficient}). We acknowledge that run times depend on implementation, though we believe a 2-3 orders of magnitude improvement to be significant, and the improvement should persist despite variations due to implementation.

% -------------------------------------------------------------

\begin{figure}[h]
\centering
\begin{subfigure}[b]{0.40\textwidth}
\includegraphics[width=\textwidth]{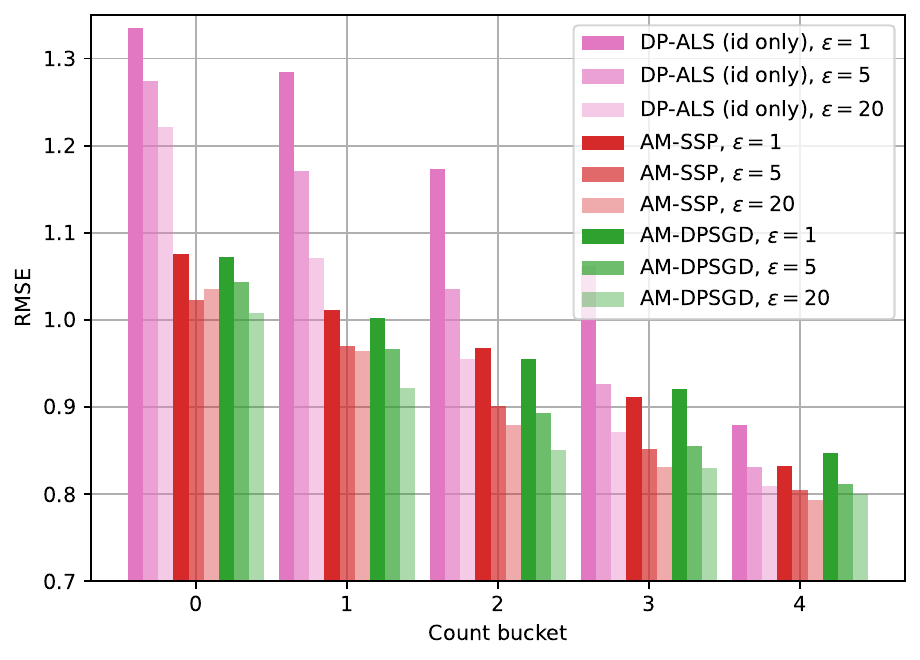}
\caption{ML10M}
\label{fig:sliced-10m}
\end{subfigure}
\hspace{.1\textwidth}
\begin{subfigure}[b]{0.40\textwidth}
\includegraphics[width=\textwidth]{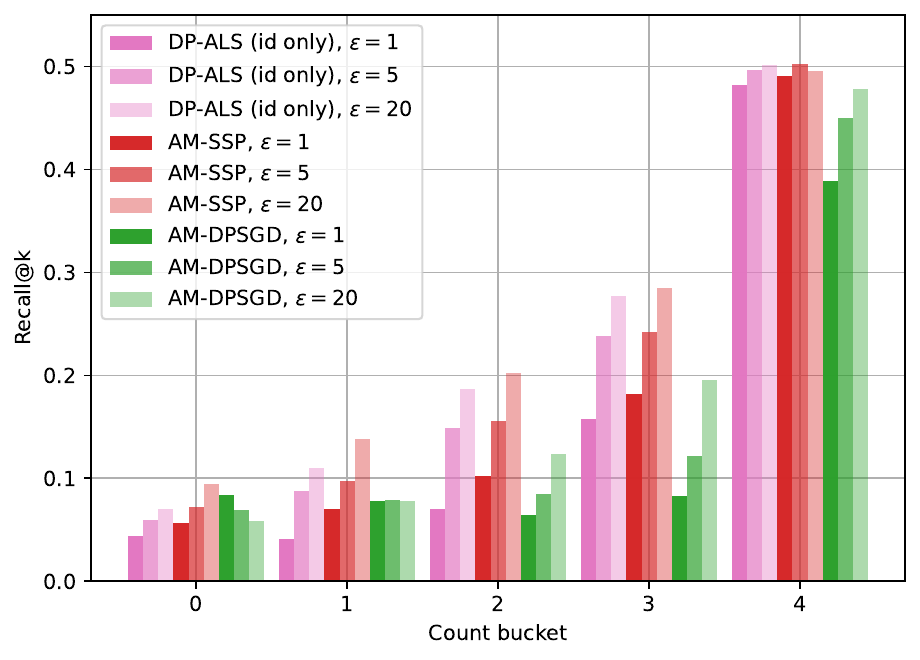}
\caption{ML20M}
\label{fig:sliced-20m}
\end{subfigure}
\caption{Sliced quality metrics. The movies are sorted by increasing count, and partitioned in five buckets of equal size. So bucket 0 corresponds to the rarest 20\% of the movies, while bucket 4 corresponds to the most frequent 20\%.}
\label{fig:sliced}
\end{figure}

\subsection{Quality on the Long Tail}
Prior work~\citep{chien2021private,krichene2023multi,curmei2023private} reported that tail movies, i.e. those with fewer training examples, are more susceptible to the noise added to guarantee DP, and this manifests in larger quality losses on the long tail. To assess the impact of our methods on tail quality, we report in Figure~\ref{fig:sliced} utility metrics sliced by movie count, for different values of $\epsilon$.

On ML10M, we observe that the AM methods perform much better on the tail than the DP-ALS method (they also improve on the top slice, but the improvement is more significant on the tail). Recall that both AM methods train a feature-based encoder, while DP-ALS trains an id based encoder. This improvement on the tail perhaps indicates that using (public) features helps learn better representations of the long tail. Now comparing AM-DPSGD and AM-SSP, is appears that AM-SSP generally performs better on the head slices, and worse on the tail slices.

On ML20M, AM-DPSGD performs worse on all slices (consistent with Figure~\ref{fig:ml20m}). Comparing AM-SSP to DP-ALS, we do observe more improvements on tail slices.

% -------------------------------------------------------------
\section{AM-SSP in Federated Learning}
\begin{figure}[h!]
\centering
\includegraphics[width=.9\textwidth]{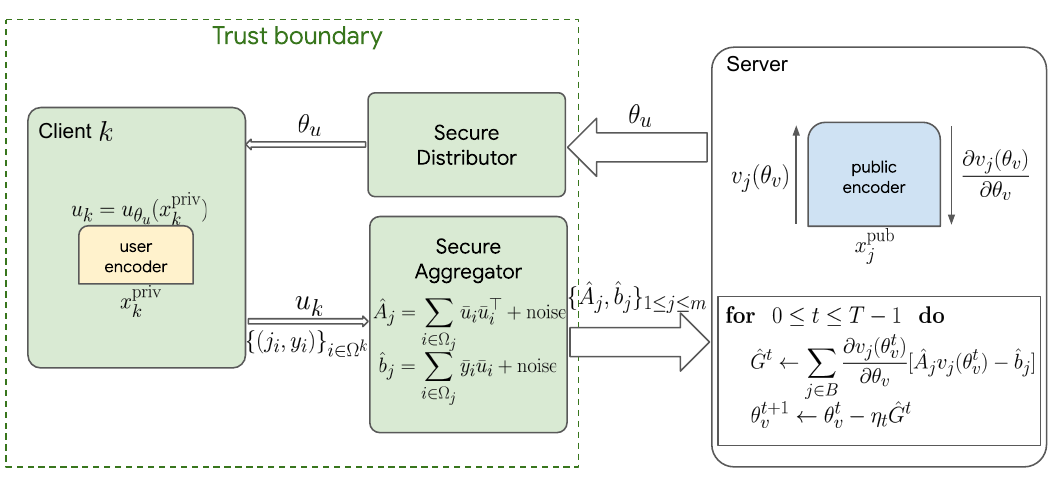}
\caption{Illustration of Federated SSP2}
\label{fig:fl}
\end{figure}

Alternating minimization has also been studied in the Federated Learning (FL) literature~\citep{singhal2021federated,collins2021exploiting}, where the goal is to perform distributed training while keeping each user's data on the user's device. The class of multi-encoder models is particularly well-suited to the federated setting, as the model reflects the natural separation between the user's data (to be protected) and the public features. In this section, we discuss implications of our algorithms on the FL setting. Figure~\ref{fig:fl} is an illustration of a federated implementation of the SSP2 algorithm.

The system consists of the following components:
\begin{itemize}[leftmargin=13pt,topsep=0pt,itemsep=0ex,partopsep=0ex,parsep=1pt]
\item A server that holds the public feature matrix $\Xpub$. The server is assumed untrusted (so any data or model parameters sent to the server need to be DP protected).
\item $n$ client devices (one per user). Client $k$ holds the training data of user $k$, consisting of the private feature vector $\xpriv_k$, together with the items and labels of the user, $\{(j_i, y_i)\}_{i \in \Ok}$.
\item A secure distributor, which simply broadcasts model parameters to the $n$ clients.
\item A secure aggregator, responsible for computing and protecting the sufficient statistics. Techniques for secure aggregation have been studied for example by~\cite{bonawitz2017secure}.
\end{itemize}

The training of the public encoder is described in Algorithm~\ref{alg:fl}. Note that the steps are identical to Algorithm~\ref{alg:mini-ssp-corr}, except that we additionally specify on which component each operation is done.

\paragraph{Advantages of SSP in the Federated Setting} Besides the advantages that we discussed in the general case (preserving gradient sparsity, noise that does not scale with the number of steps, etc.), SSP2 has additional advantages in the federated setting, compared to federated DP-SGD.

First, since we only need to add noise to the sufficient statistics once (see Remark~\ref{rem:reusing_noise}), the clients only need to be involved in \emph{one communication round}: they send data to the aggregator once, then the server takes several gradient steps. This is in contrast to federated DP-SGD, where a new communication round is initiated after each update to the encoder's parameters (the new parameters need to be broadcast to clients so gradients can be computed on clients and aggregated). Reducing the number of synchronization barriers (waiting for clients to become available) may have a significant impact on training time in practice.

\begin{algorithm}[h!]
\caption{Federated SSP2}
\label{alg:fl}
\SetAlgoNoLine
\SetKwInput{KwSend}{Send}
\DontPrintSemicolon
{\bf Distributor}\\\Indp
\KwSend{$\theta_u$ to clients.}
\Indm\vspace{.2cm}
{\bf Client $k$}\\\Indp
\KwIn{$\xpriv_k$, $\Dcal^k = \{(y_i, j_i)\}_{i \in \Ok}$}
$u_k \leftarrow u_{\theta_u}(\xpub_k)$.\\
\KwSend{$u_k, \Dcal^k$ to aggregator.}
\Indm\vspace{.2cm}
{\bf Aggregator}\\\Indp
\KwIn{Clipping parameters $\Gamma_y, \Gamma_u$, noise standard deviation $\sigma$, optional weights $\{w_i\}$}
Let $\bar u_i = \clip(u_{\theta_u}(\xpriv_i), \Gamma_u)$,
$\bar y_{i} = \clip(y_{i}, \Gamma_y)$\\
\For{$1 \leq j \leq m$}{
$\hat A_j \leftarrow \sum_{i \in \Oj} w_i\bar u_i \bar u_i^\top + \sigma\Gamma_u^2 \Ncal^{d \times d}$\\
$\hat b_j \leftarrow \sum_{i \in \Oj} w_i\bar y_i \bar u_i + \sigma\Gamma_y\Gamma_u \Ncal^d$.\\
}
\KwSend{$\{\hat A_j, \hat b_j\}_{1 \leq j \leq m}$ to server.}
\Indm\vspace{.2cm}
{\bf Server}\\\Indp
\KwIn{$\Xpub$, number of steps $T^I$, learning rate $\eta_t$, statistics $\{\hat A_j, \hat b_j\}_{1 \leq j \leq m}$}
\For{$0 \leq t \leq T^I-1$}{
Uniformly sample $B \subset [m]$\\
$\hat G^t \leftarrow \sum_{j \in B} \frac{\partial v_{\theta_v^t}(\xpub_j)}{\partial \theta_v}(\hat A_j v_{\theta_v^t}(\xpub_j) - \hat b_j)$\\
$\theta_v^{t+1} \leftarrow \theta_v^t - \eta_t \hat G^t$
}
\KwRet $\theta_v^T$
\end{algorithm}%
\vspace{.2in}%

Second, the communication cost is drastically reduced. Each client only needs to send to the aggregator the output of the private encoder ($u_k$), and the labels $\{j_i, y_i\}_{i \in \Ok}$. Contrast this with federated DP-SGD, where client need to send model \emph{gradients} (which can be significantly larger), and they do so at \emph{each round}. In SSP2, communication cost does not scale with the number of rounds nor with the public encoder's intermediate size such as feature vocab or hidden layers (as long as its output dimension is fixed).

Finally, gradient computation can happen on an \emph{untrusted} server, since dependence on sensitive data is isolated into the protected sufficient statistics. This makes it possible to (i) reduce the computational burden on client devices (each client only needs to compute a single forward pass), and (ii) avoid having to send the public item features $\Xpub$ to the client devices. This further reduces the communication cost.

%%%%%%%%%%%%%%%%%%%%%%%%%%%%%%%%%%%%%%%%%%%%%%%%%%%%%%%%%%%%

\end{document}